\documentclass[lettersize,journal]{IEEEtran}
\usepackage{amsmath,amsfonts}
\usepackage{algorithmic}
\usepackage{algorithm}
\usepackage{array}
\usepackage[caption=false,font=normalsize,labelfont=sf,textfont=sf]{subfig}
\usepackage{textcomp}
\usepackage{stfloats}
\usepackage{url}
\usepackage{verbatim}
\usepackage{graphicx}
\usepackage{cite}
\usepackage{hyperref}
\usepackage[absolute,overlay]{textpos}
\begin{document}

\begin{textblock*}{20cm}(1.37cm,0.94cm)
\fontsize{7}{6}\selectfont
This work has been submitted to the IEEE for possible publication. Copyright may be transferred without notice, after which this version may no longer be accessible.
\end{textblock*}

\title{EmoSpeaker: One-shot Fine-grained Emotion-Controlled Talking Face Generation}

\author{Guanwen Feng, Haoran Cheng, Yunan Li*~\IEEEmembership{Member, IEEE}, Zhiyuan Ma, Chaoneng Li~\IEEEmembership{Student Member, IEEE}, Zhihao Qian, Qiguang Miao*~\IEEEmembership{Senior Member, IEEE,} Chi-Man Pun~\IEEEmembership{Senior Member, IEEE} 

        % <-this % stops a space
\thanks{Guanwen Feng, Haoran Cheng, Yunan Li, Zhiyuan Ma, Chaoneng Li, Zhihao Qian, Qiguang Miao are with the School of Computer Science and Technology, Xidian University, Xi'an 710071, China, Xi'an Key Laboratory of Big Data and Intelligent Vision, Xidian University, Xi'an 710071, China and Key Laboratory of Collaborative lntelligence Systems, Ministry of Education, Xidian University, Xi'an 710071, China. Chi-Man Pun is with the Department of Computer and Information Science, University of Macau, Macao 999078, China.(email: gwfeng\_1, zjmazy, chaonengli, zhqian\_1@stu.xidian.edu.cn, yunanli, qgmiao@xidian.edu.cn, cmpun@umac.mo.)}% <-this % stops a space
\thanks{*Yunan Li and Qiguang Miao both are corresponding authors.}}



\maketitle

% \thispagestyle{fancy}
% \fancyhead[L]{\small This work has been submitted to the IEEE for possible publication. Copyright may be transferred without notice, after which this version may no longer be accessible.}
% % \fancyhead[R]{\thepage}  % 在页眉右侧添加页码
% \renewcommand{\headrulewidth}{0pt} % 去掉页眉的横线

\begin{abstract}

Implementing fine-grained emotion control is crucial for emotion generation tasks because it enhances the expressive capability of the generative model, allowing it to accurately and comprehensively capture and express various nuanced emotional states, thereby improving the emotional quality and personalization of generated content. Generating fine-grained facial animations that accurately portray emotional expressions using only a portrait and an audio recording presents a challenge. In order to address this challenge, we propose a visual attribute-guided audio decoupler. This enables the obtention of content vectors solely related to the audio content, enhancing the stability of subsequent lip movement coefficient predictions. To achieve more precise emotional expression, we introduce a fine-grained emotion coefficient prediction module. Additionally, we propose an emotion intensity control method using a fine-grained emotion matrix. Through these, effective control over emotional expression in the generated videos and finer classification of emotion intensity are accomplished. Subsequently, a series of 3DMM coefficient generation networks are designed to predict 3D coefficients, followed by the utilization of a rendering network to generate the final video. Our experimental results demonstrate that our proposed method, EmoSpeaker, outperforms existing emotional talking face generation methods in terms of expression variation and lip synchronization. Project page: 
 \href{https://peterfanfan.github.io/EmoSpeaker/}{https://peterfanfan.github.io/EmoSpeaker/}
 
% Generating fine-grained facial animations that accurately portray emotional expressions using only a portrait and an audio recording presents a challenge. Existing methods often rely on multiple emotional portraits or a video clip to capture different emotional expressions, while some utilize emotion labels for animation generation. However, these approaches lack precise control over facial emotional expression and face issues with lip synchronization accuracy. In order to address these challenges, we propose an visual attribute-guided audio decoupler. This enables the obtention of content vectors solely related to the audio content, enhancing the stability of subsequent lip movement coefficient predictions. To achieve more precise emotional expression, we introduce a fine-grained emotion coefficient prediction module. Additionally, we propose an emotion intensity control method using a fine-grained emotion matrix. Through these, effective control over emotional expression in the generated videos and finer classification of emotion intensity are accomplished. Subsequently, a series of 3DMM coefficient generation networks are designed to predict 3D coefficients, followed by the utilization of a rendering network to generate the final video. Our experimental results demonstrate that our proposed method, EmoSpeaker, outperforms existing emotional talking face generation methods in terms of expression variation and lip synchronization. Project page: 
%  \href{https://peterfanfan.github.io/EmoSpeaker/}{https://peterfanfan.github.io/EmoSpeaker/}
\end{abstract}




\section{Introduction}
\begin{figure}[!t]\centering
	\includegraphics[width=8.5cm]{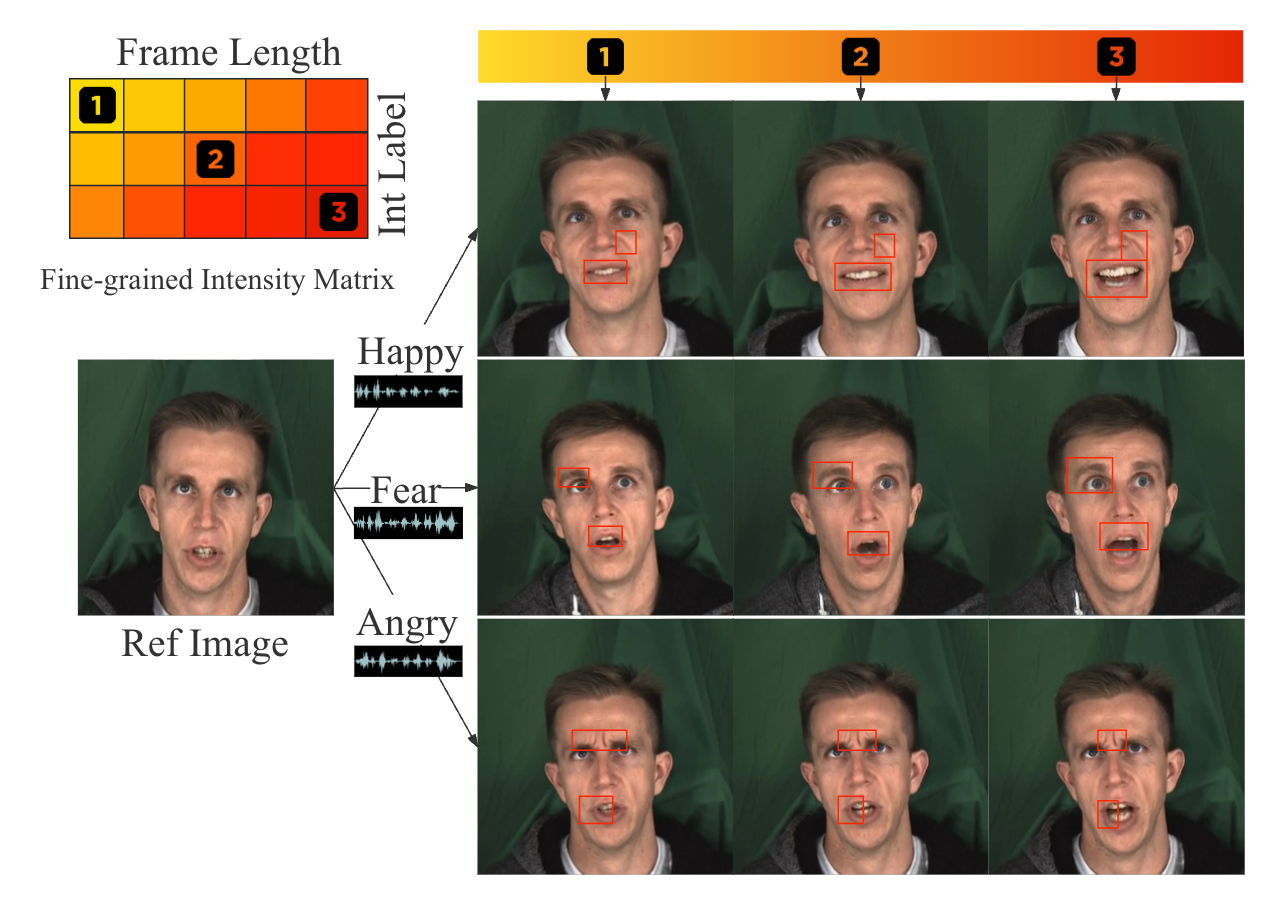}
	\caption{This is our proposed EmoSpeaker that generates speaker videos from a single image, driving audio, and specifying emotion category labels and fine-grained emotion intensity. We have marked the variation of different AU regions in different fine-grained.}
    \label{zhaiyao}
\end{figure}
Audio-driven talking face generation has become a research focus in recent years. This technology has various application scenarios, such as virtual digital human generation \cite{wang2021adversarial,yu2020human}, virtual reality \cite{zhu2021learning,li2020interactive} and movie special effects \cite{henderson2021digital,liu2021high}. Previous research has paid more attention to improve lip-synchronization \cite{prajwal2020lip} and video generation quality, but there has been limited exploration of the emotional expression of generated videos. Some current studies have addressed emotional facial animation generation \cite{karras2017audio,livingstone2018ryerson,sadoughi2019speech,wang2020mead,ji2021audio,ji2022eamm}. 

% 下面这段扩充
Recent emotional face generation methods are always driven by either long or short video \cite{edwards2016jali,karras2017audio,li2021write}.
Using a label-controlled approach, it is difficult to generate emotional videos of different intensities with different emotional intermediate states \cite{wang2020mead, ji2021audio,ji2022eamm}. The most one-shot generation methods generally consider only lip-synchronicity without consideration of emotional factors \cite{wang2021one,wang2022one,yin2022styleheat,zhang2023sadtalker,ma2023otavatar}. 
% 加点one-shot的方法

Audio usually contains speaker's emotion information and linguistic content information, with emotion mainly controlling  expression and linguistic content controlling lip movements. In order to generate emotional faces with arbitrary granularity and highly synchronised lip movements, it is necessary to separate the emotion vectors and content vectors in audio. This requires the design of a highly reliable emotion decoupling method as well as an emotion injection method with variable and smooth intensity control. 

To address the problems of lip-synchronisation as well as fine-grained intensity control and generation of arbitrary intensity emotions on arbitrary faces, we propose the EmoSpeaker method, which is driven by 3D coefficients as intermediate representations to bridge different parts of the talking face generation process.
% 第一个模块
% 加一下对比学习的参考文献
To achieve this goal, we first introduce Visual Attribute-Guided Audio Decoupler. In this module, our aim is to extract the content vector from the audio without mixing emotional information. By utilizing the AU-based Contrastive Learning \cite{chen2020simple} to guide the audio encoder, we decouple the emotion vector from the audio clip and obtain pure content feature in the audio for more accurate control of the lip motion. Next, in Fine-grained Emotion Coefficient Prediction Module, the content vectors are aggregated with the target emotion category and emotion intensity information corresponding to the training data. Furthermore, we explore a fine-grained intensity control scheme that generates videos with unseen emotion intensity expressions beyond the train dataset domain. 
% The prediction of 3D coefficients is accomplished using ExpNet, EmoNet, and PoseNet.
% 第三个模块
Finally, in Emotion Face Renderer, we design a mapping network to map the generated 3D face model coefficient to the motion parameters of potential key points. These motion parameters are used to drive the facial motion of the potential key points in the reference image, resulting in the generation of the final video.

Our core contributions are as follows:
% The contributions of this paper can be summarized as follows: 
\begin{itemize}
\item 
We propose a one-shot fine-grained emotion-controlled talking face generation method EmoSpeaker for generating highly realistic speaker videos with the ability to artificially control emotional categories and fine-grained emotional intensity while achieving the precise lip-synchronisation.
\item 
We develop a visual attribute-guided audio decoupler. This aims to exploit the correlation between facial AUs and emotional expressions by removing confounding emotion vectors from content vectors.
\item 
We develop a fine-grained emotion intensity control module. The fine-grained emotion representation is accomplished in a more detailed and accurate way by specifying the emotion categories and the emotion intensities in the fine-grained emotion intensity matrix.
\end{itemize}

\section{Related work}
 
\begin{figure*}[t]
	\centering
	\includegraphics[width= 1.0\textwidth]{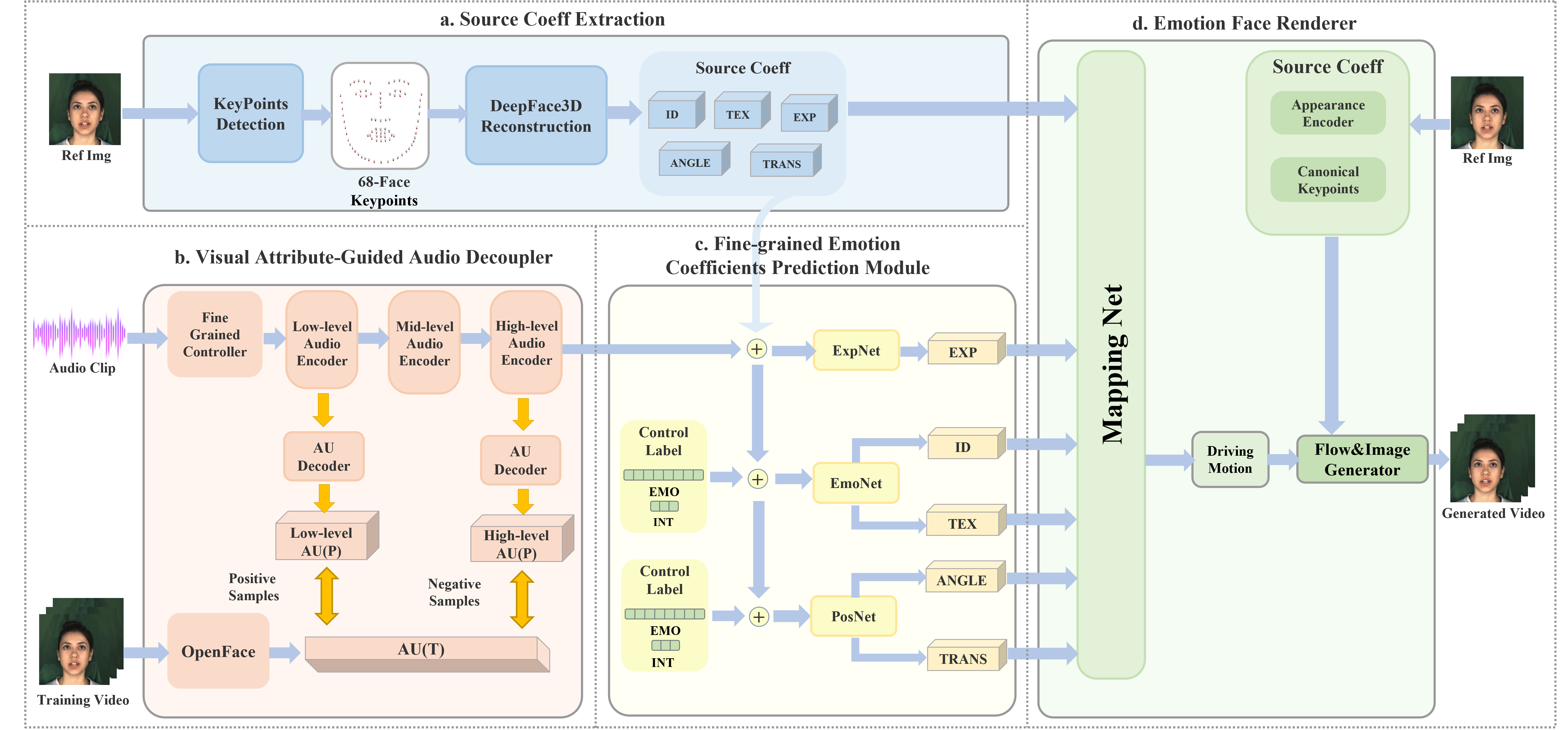} % Reduce the figure size so that it is slightly narrower than the column.

        \caption{We perform multi-level encoding on the driving audio to obtain content vectors. These vectors are then combined with coefficients from reference images, emotion type labels, and emotion intensity labels. The predicted coefficients are obtained. Finally, using MappingNet, we map these coefficients to driving motion coefficients, distort the reference images, and generate the final video. \textbf{a. Source Coeff Extraction}: Extract 68 facial keypoints and 3DMM coefficients from the reference image for training and generation purposes.  \textbf{b.Visual Attribute-Guided Audio Decoupler}: Input the audio into three consecutive audio encoders to obtain separate low-level and high-level audio encodings.   Utilizing a shared AU decoder to obtain AU-related features and compare them with AU coefficients extracted from the training videos for comparative learning.  \textbf{c.Fine-grained Emotion Coefficient Prediction Module}: Manually specify emotion categories and intensity labels.   During inference, adjust the sliding window size of the input audio to obtain a fine-grained emotion vector synchronized with the audio.   Combine them with content vectors to predict expression, emotion, and pose coefficients through ExpNet, EmoNet and PoseNet.  \textbf{d. Emotion Face Renderer}: Utilize the predicted 3DMM coefficients to generate motion vectors for latent facial keypoints, animating the facial image.}
	\label{overview}
\end{figure*}

\begin{figure*}[t]
	\centering
	\includegraphics[width= 0.9\textwidth]{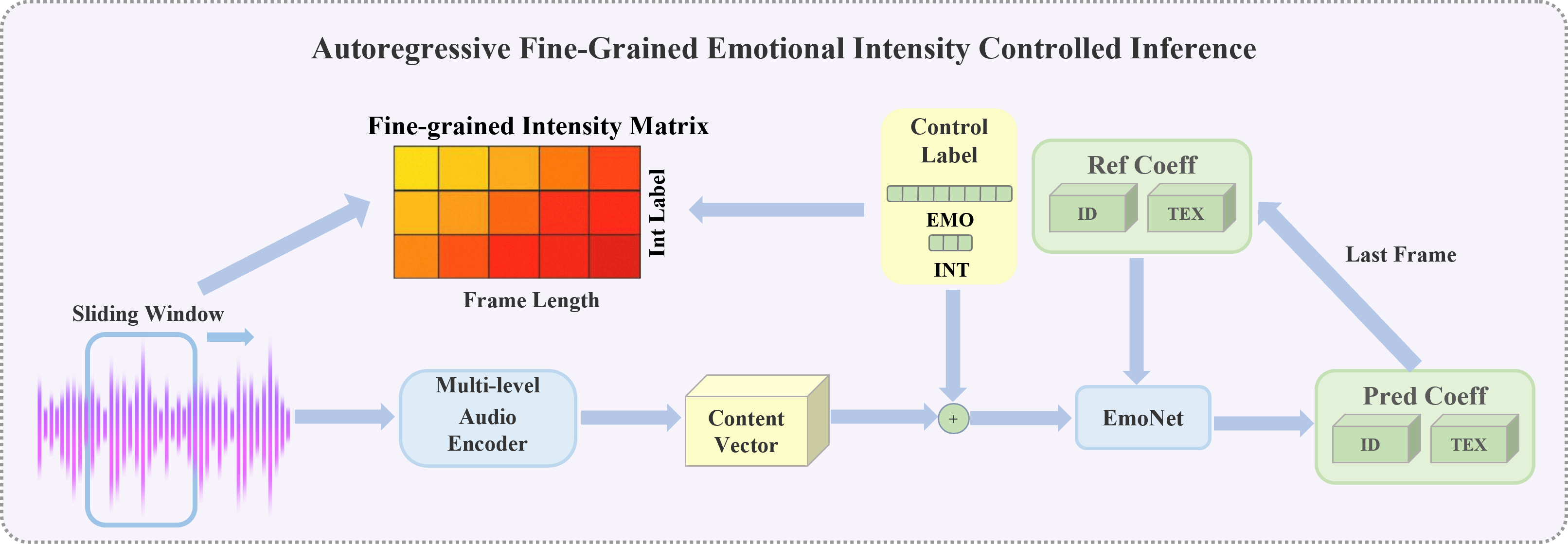} % Reduce the figure size so that it is slightly narrower than the column.
    	\caption{The flowchart of Fine-grained Emotion Coefficient Prediction. Audio sliding windows of varying sizes are utilized during the inference process. Different emotion categories and intensity labels are manually assigned. The predicted coefficients is obtained by EmoNet. Subsequently, the predicted coefficients of the last frame serve as the reference coefficients for the consecutive window.}
	\label{fg}
\end{figure*}

\begin{figure*}[t]
	\centering
	\includegraphics[width= 0.9\textwidth]{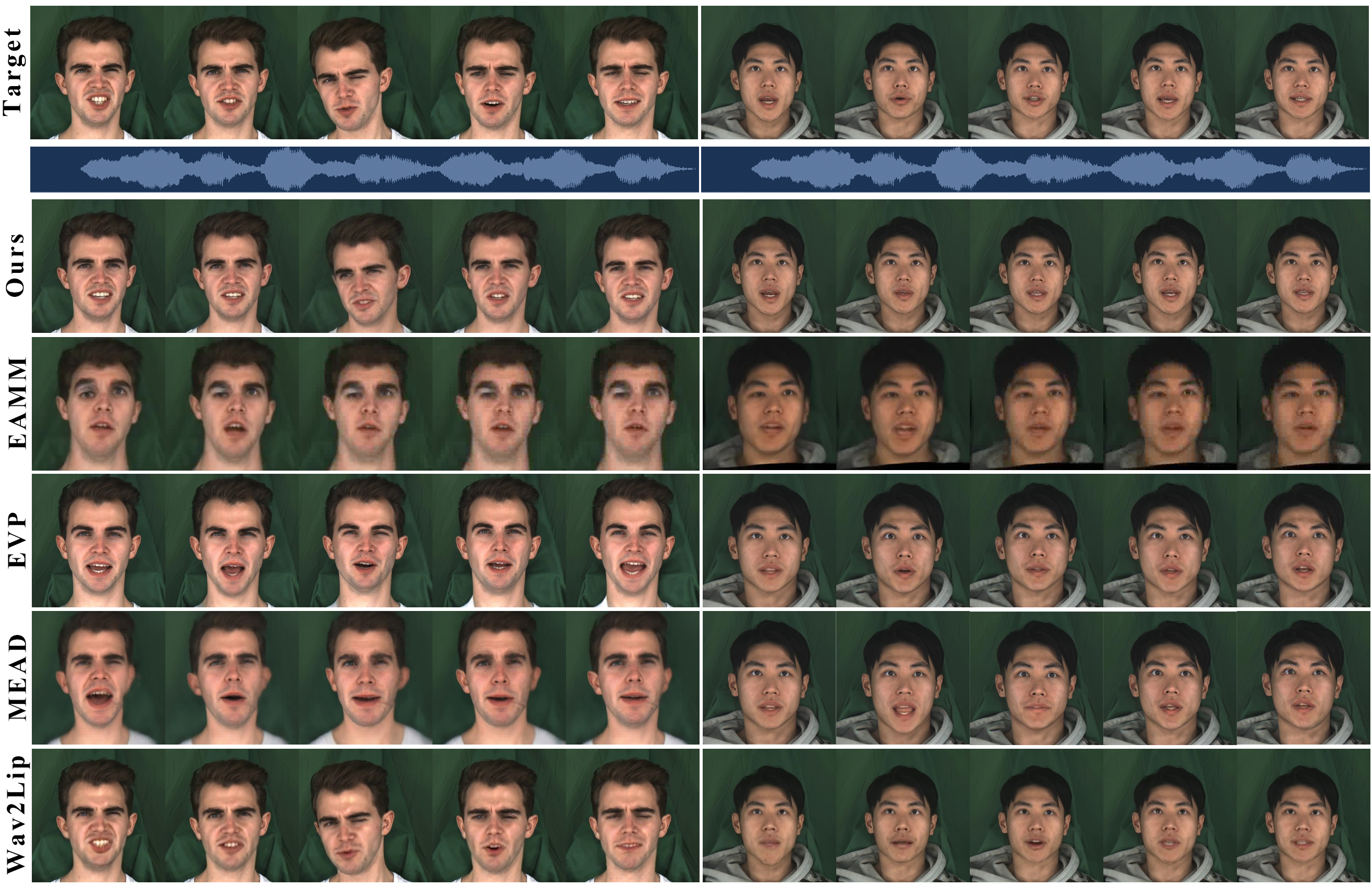} % Reduce the figure size so that it is slightly narrower than the column.
    	\caption{ We compare our method with the state-of-the-art emotion-driven facial expression generation methods such as EAMM, EVP, and MEAD, as well as the lip generation method Wav2lip. It is evident from the figure that our method exhibits superiority in various aspects including lip synchronization, pose reconstruction, and video quality.}
	\label{duibi}
\end{figure*}

\subsection{Audio-driven Talking Face Generation}
Audio-driven facial animation generation is a technique that use audio input to generate facial animation or expressions that correspond  the audio content. It combines techniques like audio processing, facial expression analysis, and animation synthesis to achieve more concise and natural facial expressions. In generating videos with facial animation, the fundamental challenge is to generate mouth movements that are synchronized with speech content, with a primary focus on modeling the mouth region \cite{suwajanakorn2017synthesizing,wang2012high}. In particular, the \cite{prajwal2020lip} method generates precise mouth shapes for individuals, significantly enhances \textbf{Lip Synchronization}, and offers a robust pre-training model. Subsequently, facial animation generation can be broadly categorized into \textbf{Person-specific} and \textbf{Person-independent} methods. Some methods employ GAN networks to create video versions specific to individuals, implicitly incorporating visual information \cite{zhang2021facial,sun2021speech2talking}, including facial details and background, into the network. Speaker-independent approaches are implemented using various algorithmic frameworks \cite{wang2022one,doukas2021headgan,zhou2021pose,zhou2020makelttalk,DBLP:conf/ijcai/WangLDFY21,zhang2023sadtalker,ren2021pirenderer,zhang2021flow}. Some previous methods lack consideration of  speaker information  and head movement. \cite{zhou2020makelttalk} extracts speaker information from speaker audio clips to control facial dynamics. \cite{zhou2021pose} captures the pose pattern of the driver video to apply to the target video. \cite{zhang2023sadtalker} and \cite{ren2021pirenderer}utilize 3D coefficient representations for one-shot generation, but  none of these methods take into account emotional factors.

\subsection{Emotional Talking Face Generation}
Emotion plays a crucial role in facial expression, enhancing both the attractiveness and expressiveness of facial animation.  In speech-driven face generation, the emotion information extracted from the audio needs to be mapped to the corresponding facial expressions or actions. This mapping can be achieved through the use of deep neural networks that model the correspondence between audio emotion and facial expression. \cite{sadoughi2019speech} focuses on learning the correlation between emotion and lip movement. \cite{wang2020mead} collect the MEAD dataset, encode different emotional categories in one-hot manner, and control the generated emotional video categories based on specified emotional labels. \cite{ji2021audio} propose a method for separating speech decomposition into content encoding and emotion encoding using cross-reconstruction, allowing for the synthesis of emotion dynamics from audio. These methods are only applicable to portraits that were included in the training dataset and cannot generate videos of individuals who were not seen during training. \cite{sinhaemotion2022ijcai} selects a
limited range of emotion intensities from the dataset but allows for finer emotion control, enabling generation of emotion levels beyond dataset labels. In contrast, \cite{ji2022eamm} separates emotional control in a one-shot mode but lacks control over emotional intensity. Our method, on the other hand, enables fine-grained control over emotional intensity for any individual in a one-shot manner.

\subsection{Video-driven Talking Face Generation}
Video-driven talking face generation, also known as facial reenactment, is closely related to audio-driven facial animation generation. These approaches involve transferring motion patterns from audio or video to the target person. Previous approaches are mainly categorized into those based on 2D landmark, \cite{ji2021audio,DBLP:conf/nips/Wang0TLCK19} 3DMM \cite{doukas2021headgan,ren2021pirenderer,ma2023otavatar,zhang2023sadtalker,yin2022styleheat}, motion field \cite{zakharov2020fast,siarohin2019first,siarohin2019animating} and feature decoupling \cite{burkov2020neural}.
Our method is primarily based on 3DMM, and in the video rendering step, we draw inspiration from \cite{wang2021one} and video rendering network proposed by \cite{zhang2023sadtalker}. The final video is generated by using the rendering network to combine the set of 3D coefficients predicted in the previous steps with the reference image.

\section{Methodology}

\begin{algorithm}
\caption{EmoSpeaker}
\label{algo}
\textbf{Input:} audio segment $A$, synchronized image sequence $Q$, emotional category label $E\{angry, xxx, \}$, specified emotional intensity $P$

\textbf{Output:} target video

\vspace{\baselineskip}

\textbf{Preprocessing}

1. Determine audio sliding window size $t$ and emotional intensity label $Pl$ based on selected emotional intensity $P$

2. Extract corresponding Mel spectrogram from the input audio:

\hspace{0.5cm} $M(m_1, m_2, ..., m_i) = M(A)$

3. Extract corresponding Action Unit (AU) coefficients and 3DMM coefficients from the input images:

\hspace{0.5cm} $AU(au_1, au_2, ..., au_i) = O(Q)$

\hspace{0.5cm} $C(c_1, c_2, ..., c_i) = D(Q)$

\hspace{1cm} where $c_i$ represents five coefficients: $id_i$, $tex_i$, $exp_i$, $angle_i$, $trans_i$

% \vspace{\baselineskip}

\textbf{Begin}

1. \textbf{for} $m_i$, $au_i$, $c_i$ in $M$:

2. \hspace{0.5cm} // AU contrastive learning to remove emotional features from audio

3. \hspace{0.5cm} $feat\_low_i$, $feat\_high_i$, $content\_vector_i$ 

    \hspace{1.5cm} = AudioEncoder($m_i$)

4. \hspace{0.5cm} $au\_low_i$, $au\_high_i$, $content\_vector_i$

    \hspace{1.5cm} = AudioDecoder($m_i$)

5. \hspace{0.5cm} // Compute contrastive loss

6. \hspace{0.5cm} $\mathcal{L}_{cross}$ = $f_{contras}$($au\_low_i$, $au\_high_i$, $au_i$)

7. \hspace{0.5cm} // Coefficient prediction

8. \hspace{0.5cm} $exp\_pred_i$ = ExpNet($content\_vector_i, exp_{ref}$)

9. \hspace{0.5cm} $id\_pred_i$, $tex\_pred_i$ = EmoNet($content\_vector_i$, $id_{ref}$, $tex_{ref}$, $E$, $P$)

10. \hspace{0.5cm} $angle\_pred_i$, $trans\_pred_i$

   \hspace{1cm} = PoseNet($content\_vector_i$, $angle_{ref}$, $trans_{ref}$, $E$, $P$)

11. \hspace{0.5cm} // Compute loss

12. \hspace{0.5cm} $\mathcal{L}_{pred}$ = $f_{l1}$($c_i$, $c\_pred_i$) + $f_{cls}$($id\_pred_i$, $tex\_pred_i$)

% 12. \hspace{0.5cm} $$\(\mathcal{L}_{pred} = f_{l1}(c_i, c\_pred_i) + f_{cls}(id\_pred_i, tex\_pred_i)\)$$

13. \textbf{endfor}

% \vspace{\baselineskip}

14. // Generate video

15. \textbf{for} $c\_pred_i$ in $C_{pred}$:

16. \hspace{0.5cm} $\mathcal{L}_{map}$ = $f_{l1}$($k_i$, $k\_pred_i$)   // k is a potential key point in Face-vid2vid

17. \hspace{0.5cm} $Ipred_i$ = $EmoFaceRenderer$(MappingNet($c\_pred_i$))
% //TODO（添加损失函数）

18. \textbf{endfor}

\end{algorithm}

We propose a one-shot fine-grained emotion-controlled talking face generation method EmoSpeaker shown in Figure \ref{zhaiyao}. The model generates a video of the target's face by inputting a reference image, an  audio clip, a specified emotion category, and fine-grained intensity. The overall framework is shown in Figure \ref{overview}. The  model consists of three main components Visual Attribute-Guided Audio Decoupler, Fine-grained Emotion Coefficient Prediction Module and Emotion Face Renderer. The detailed analysis of preliminary  and  each module will be provided in subsequent chapters.

% As show in \ref{overview} 
% \begin{figure*}[t]
% \centering
% \includegraphics[width=0.8\textwidth]{figure2} % Reduce the figure size so that it is slightly narrower than the column.
% \caption{Adjusting the bounding box instead of actually removing the unwanted data resulted multiple layers in this paper. It also needlessly increased the PDF size. In this case, the size of the unwanted layer doubled the paper's size, and produced the following surprising results in final production. Crop your figures properly in a graphics program. Don't just alter the bounding box.}
% \label{fig2}
% \end{figure*}

\subsection{Preliminary}
\subsubsection{3D Morphable Model}
% Our approach uses 3DMM coefficients as an intermediate representation for talking head generation. I first extract the 3DMM coefficients of the face in the reference frame from the reference image, and after combining it with the audio, I predict the corresponding audio-synchronized face coefficients using three networks: expnet, idtexnet (emonet), and posenet.

% 3DMM \cite{blanz1999morphable} is a statistical model for modeling and generating face shape and texture. It is trained on a large dataset of 3D faces and can be parametrized to represent variations in facial features.A 3DMM model typically consists of two main components: a shape model and a texture model. The shape model describes the geometric structure of the face, including keypoints, contours, and curves. The texture model, on the other hand, captures the color and texture information of the face surface. By adjusting the parameters of the 3DMM model, control of the shape and texture of the face can be realized. These parameters can be represented as linear combinations, thus allowing interpolation and transformation in shape and texture space. This makes the 3DMM model very useful in applications such as face reconstruction, expression synthesis and face animation.The 3DMM model can be expressed as:

3DMM \cite{blanz1999morphable} is a statistical model for modelling and generating face shapes and textures. It is trained on a large 3D face dataset and can be parameterised to represent changes in facial features. The model consists of two main parts: a shape model and a texture model. The shape model describes the geometric structure of the face, while the texture model captures the colour and texture information of the face surface. The control of the shape and texture of the face can be achieved by adjusting the parameters of the 3DMM model, which can be expressed in Eq \ref{Eq:3dmm}:
\begin{equation}
S=\bar{S}+W_{\text {id }} U_{i d}+W_{\text {tex }} U_{\text {tex }}+W_{\text {exp }} U_{\text {exp }}
% S=\bar{S}  + W_{id}U_{id}+ W_{tex}U_{tex}+ W_{exp}U_{exp}+ W_{angle}U_{angle}+ W_{trans}U_{trans}
\label{Eq:3dmm}
\end{equation}
where $U_{i}$ denotes different feature vectors and $W_{i}$ denotes the corresponding weights, specifically, $U_{id}\in \mathbb {R}^{80}$  , $U_{tex}\in \mathbb {R}^{80}$ corresponds to the identity and texture features of different face models, and $U_{exp}\in \mathbb {R}^{64}$ denotes the expression coefficients of the face. In addition, in order to control the movements of the face, we also utilize $P_{angle}\in \mathbb {R}^{3}$,  $P_{trans}\in \mathbb {R}^{3}$ to denote the pose coefficients of the face.

\begin{table}[t]
  \centering
  \caption{Facial Action Units (AU) Information}
  \label{tab:au_info}
  \resizebox{0.48\textwidth}{!}{%
\begin{tabular}{|c|c|c|}
    \hline
    \textbf{AU} & \textbf{Description} & \textbf{Movement Degree} \\
    \hline
    AU1 & Inner Brow Raiser & Upward movement of the inner brow \\
    \hline
    AU2 & Outer Brow Raiser & Upward movement of the outer brow \\
    \hline
    AU4 & Brow Lowerer & Downward movement of the brow \\
    \hline
    AU5 & Upper Lid Raiser & Upward movement of the upper lid \\
    \hline
    AU6 & Cheek Raiser & Upward movement of the cheekbone \\
    \hline
    AU7 & Lid Tightener & Tightening of the eyelid \\
    \hline
    AU9 & Nose Wrinkler & Wrinkling of the nose \\
    \hline
    AU10 & Upper Lip Raiser & Upward movement of the upper lip \\
    \hline
    AU12 & Lip Corner Puller & Upward movement of the lip corners \\
    \hline
    AU14 & Dimpler & Dimpling of the cheek \\
    \hline
    AU15 & Lip Corner Depressor & Downward movement of the lip corners \\
    \hline
    AU17 & Chin Raiser & Upward movement of the chin \\
    \hline
    AU20 & Lip Stretcher & Horizontal stretching and flattening of lips \\
    \hline
    AU23 & Lip Tightener & Tightening of the lips \\
    \hline
    AU25 & Lips Part & Parting of the lips \\
    \hline
    AU26 & Jaw Drop & Downward movement of the jaw \\
    \hline
    AU45 & Blink & Frequency of blinking \\
    \hline
  \end{tabular}%
	}
\end{table}

\subsubsection{Facial visual representation}
Facial Action Units (AUs) play a crucial role in the study of facial expressions \cite{ekman1978facial}. AUs are designed to describe the movement patterns of facial muscles in Table \ref{tab:au_info}, and there is a close association between the movement of facial muscles and human emotions and expressions. Different AUs correspond to different facial expressions, and they can be used individually or in combination to express various emotions and emotional states as shown in Table \ref{tab:emotion_au}. By observing and comparing the movement patterns of multiple AUs, it is possible to more accurately assess a person's emotional state.
\begin{table}[!htb]
  \centering
  \caption{Facial Expression AUs and Emotions}
  \label{tab:emotion_au}
  \begin{tabular}{|c|c|}
    \hline
    \textbf{Emotion} & \textbf{Facial Action Units (AU)} \\
    \hline
    \textbf{Anger} & AU4, AU5, AU7, AU23 \\
    \hline
    \textbf{Contempt} & AU12, AU14 \\
    \hline
    \textbf{Disappointment} & AU1, AU15 \\
    \hline
    \textbf{Fear} & AU1, AU2, AU5, AU25 \\
    \hline
    \textbf{Sadness} & AU1, AU4, AU15 \\
    \hline
    \textbf{Calm} & None specific \\
    \hline
    \textbf{Surprise} & AU1, AU2, AU5, AU26 \\
    \hline
    \textbf{Happiness} & AU6, AU12, AU25 \\
    \hline
  \end{tabular}
\end{table}

\subsection{Visual Attribute-Guided Audio Decoupler}

Due to the inherent differences between audio and text, audio features encompass both content and emotion features. Therefore, accurately predicting lip information directly from raw audio features is challenging. To address this issue, it is crucial to separate content and emotion features and utilize content features highly correlated with lip movements to predict 3D coefficients. However, decoupling these two feature vectors solely from a single audio reference is difficult. Previous experiments we conducted using existing excellent audio feature extractors failed to yield accurate results \cite{amodei2016deep,baevski2020wav2vec}.

Hence, we propose a Visual Attribute-Guided Audio Decoupler. The core idea of this module is to leverage visual information, such as the motion patterns of Facial Action Units (AU), to guide the decoupling process of audio. Visual information guidance enhances the precision and controllability of decoupling, as facial actions are typically associated with the speaker's emotional state and expression. In our approach, we use AUs as visual information to guide the emotion decoupling process. By employing AUs as visual information, we can understand the motion patterns of facial expressions deeply, thereby improving control over the emotional expression in the driven audio.

This decoupler provides greater flexibility, allowing the system to independently handle the content and emotion of speech and offering more adjustment and customization possibilities for subsequent processing steps. Specifically, our approach employs multi-level audio encoders, divided into low, medium, and high levels. The low-level encoder captures information closely related to emotion and content, while the high-level encoder's output aligns more with speech content information. Our aim is to extract a pure content vector from the high-level speech encoding. We use the same AU decoder to map low-level and high-level speech features to low-level AU information and high-level AU information in Eq \ref{eq:fau}. AU information extracted from the training dataset is utilized to construct a contrastive loss for generating positive samples against using low-level AU information in Eq \ref{eq:loss_contras} . 
% By introducing such a module, the system can better understand and differentiate between emotional and content features. 
% Compared to existing audio encoders, the Visual Attribute-Guided Audio Decoupler is more suitable for this task.

\begin{equation}
f_{au}^l=\boldsymbol{D}_{au}(\boldsymbol E_a^l(a_t))
\label{eq:fau}
\end{equation}

% l∈{low、mid、high}
% Dau是au解码器
% Eal是第l层音频编码器

\begin{equation}
\mathcal{L}_{\text {contras }}=-\log \frac{\exp \left(\operatorname{sim}\left(f_{a u}, f_{a u}^{\text {low }}\right) / \tau\right)}{\sum_{i=1}^K \sum_{l \in L} \exp \left(\operatorname{sim}\left(f_{a u}, f_{a u(i)}^l\right) / \tau\right)}
\label{eq:loss_contras}
\end{equation}
where $f_{a u}$ represents AU feature. $a_t$ represents the Mel-Frequency Cepstral oefficients (MFCC) features of an audio frame at time t. $E_a^l$ is the audio encoder for the l-th layer and ${D}_{au}$ is the AU decoder. $\mathcal{L}_{\text {contras }}$ is contrastive Loss.  $L = $ \{low, mid, high\}.

During the training process, the emotional vector is removed from the audio vectors, ultimately yielding content vectors related to lip movement from the high-level speech encoding.
\subsection{Fine-grained Emotion Coefficient Prediction Module}
After acquiring pure content vectors that are strongly associated with lip movements, we develop an ExpNet to convert these content vectors into Exp coefficients in Eq \ref{eq:loss_exp}, which control the lip movements. 
\begin{equation}
  \mathcal{L}_{\text {exp}}=||\boldsymbol{D}_{ExpNet}(\boldsymbol E_a(a_t)) - \hat {Exp}_t ||_2
\label{eq:loss_exp}
\end{equation}
where $ \mathcal{L}_{\text {exp}}$ is Exp coefficient loss in Eq \ref{eq:loss_exp}. ${D}_{ExpNet}$  is ExpNet. $E_a^l$ is the AudioEncoder. $a_t$ denotes the audio segment at time t, and ${Exp}_t$ represent actual Exp value.

To produce videos with specific emotions, we also explore the integration of artificial emotional vectors into the content vectors. This involves considering both the emotion category and intensity. We introduce a fine-grained emotion intensity matrix shown in Figure \ref{fg} that incorporates intensity labels and audio sliding windows. 

\textbf{ Fine-grained emotion intensity matrix: } During training, we employed the same audio window size and various combinations of emotion intensity labels to generated videos. These emotion intensity labels influence each frame of audio within the audio window, and through training, the cumulative impact of emotion intensity labels on the frames within the same audio window is determined. Consequently, during the inference stage, adjusting the size of the audio window magnifies or diminishes the influence on each frame of audio, thereby affecting the emotional intensity contained in each audio frame. Specifically, we treat the audio window size as one of the control variables for emotion intensity. By combining the audio window size and emotion intensity labels during the inference stage, we obtain emotional intensity results that cannot be achieved with a single emotion intensity label alone. We refer to this combination as a fine-grained emotion intensity matrix. 

Furthermore, we propose an EmoNet to predict Id and Tex coefficients. 
\begin{equation}
\mathcal{L}_{\text {emo}}=||\boldsymbol{D}_{EmoNet}(\boldsymbol E_a(a_t),e,p) - [\hat{Id},\hat{Tex}]_t ||_2 
\label{eq:loss_emo}
\end{equation}
where $ \mathcal{L}_{\text {emo}}$ is emotional coefficient loss in Eq \ref{eq:loss_emo}. ${D}_{EmoNet}$ is EmoNet. $e$ is emotional category, and $p$ is intensity label. $[\hat{Id},\hat{Tex}]$ represent actual Id and Tex value.

\begin{equation}
    \mathcal{L}_{\text {cls}}=-\sum e\log{\boldsymbol C_e({[Id,Tex]})}
\label{eq:loss_cls}
\end{equation}
where $ \mathcal{L}_{\text {cls}}$ is emotional coefficient classification loss in Eq \ref{eq:loss_cls}. $ C_e$ is emotional coefficient classification network. 

% By specifying various sizes of audio sliding windows during inference, we could repeatedly incorporate emotion label information within the same audio length, resulting in an overlay effect. Moreover, due to the variability in sliding window sizes, the impact of emotion intensity labels on audio encoding also varies, ultimately leading to the generation of videos with fine-grained emotion intensities. 
\begin{equation}
\mathcal{L}_{\text {pose}}=||\boldsymbol{D}_{PoseNet}(\boldsymbol E_a(a_t),e,p) -  [\hat{Angle},\hat{Trans}]_t ||_2
\label{eq:loss_pose}
\end{equation}
where $ \mathcal{L}_{\text {pose}}$ is pose coefficient loss in Eq \ref{eq:loss_pose}. ${D}_{PoseNet}$ is PoseNet. $e$ is emotional category, and p is intensity label. $[\hat{Angle},\hat{Trans}]$ represent actual Angle and Trans value. 
Similarly, to generate actions that are more aligned with the emotions, we design a PoseNet that predicts the corresponding Angle and Trans action coefficients based on the input content vectors and specified emotional information.

\begin{equation}
    \mathcal{L_{\text {total}}}= \alpha \mathcal{L}_{contras} + \beta\mathcal{L}_{exp} + \gamma \mathcal{L}_{emo} + \delta \mathcal{L}_{cls} +  \epsilon \mathcal{L}_{pose}
\label{eq:Loss_total}
\end{equation}
$\mathcal{L_{\text {total}}}$ is total loss in \ref{eq:Loss_total}. $\alpha$, $\gamma$, $\delta$ and $\epsilon$ represent different weights and are set to 5,5,3,1,1. As part of the task of generating speaker videos, we prioritize ensuring lip-audio synchronization. Therefore, we assign high weights to the training losses for the audio encoder: $\mathcal{L}_{contras}$ and $\mathcal{L}_{exp}$. Next, we focus on the fusion of emotional information. Since emotional categories are easier to distinguish, but it's challenging to control the intensity of fine-grained emotions, we assign a weight to $\mathcal{L}_{emo}$ that is second only to the lip loss. Meanwhile, the loss $\mathcal{L}_{cls}$, which aids in the training of emotional categories, is given a lower weight. Additionally, as the head poses in the training set mostly remain unchanged, we do not need to pay much attention to changes in pose, assigning a low weight to $\mathcal{L}_{pose}$.
\subsection{Emotion Face Renderer}
% After generating realistic 3D face coefficients, we need to utilize these coefficients to drive the faces in the reference images to perform corresponding actions. For this purpose, we need to train an image animation renderer.  Our approach is influenced by Face-vid2vid, which effectively establishes the relationship between potential keypoints and the motion of pixel points in real face images through implicit modeling of facial keypoints within single images. During the inference stage, Face-vid2vid leverages potential keypoints from driver images to drive corresponding motions in the reference image. Building upon the pre-trained Face-vid2vid model, we utilize the obtained multiple 3DMM coefficients (Id, Tex, Exp, Angle, and Trans), with each coefficient controlling specific information. These coefficients are fed into a mapping network within the Face-vid2vid framework to derive the necessary potential keypoint locations in the driver image. Subsequently, rendering is performed to generate the final video output.

After generating real 3D facial coefficients, we need to utilize these coefficients to guide the face in the reference image to perform corresponding actions. To achieve this goal, it is essential to train an image animation renderer. Our approach is influenced by Face-vid2vid, a method that implicitly models facial keypoints in individual images, effectively establishing the relationship between latent keypoints in real facial images and pixel movements. During the inference phase, Face-vid2vid leverages latent keypoints from the driving image to manipulate the corresponding movements in the reference image. We design a MappingNet that takes the predicted 3DMM coefficients as input. Through the predictions of MappingNet, we obtain motion coefficients for latent facial keypoints, and subsequently generate videos using the Flow\&Image Generator.
\begin{equation}
    \mathcal L_{m} = ||M([c_i])-k_i||_2
\end{equation}
 where $\mathit{M}$ represents MappingNet, $\mathit{ci}$ represents 3DMM coefficients, with $C_i \in \{Id, Tex, Exp, Angle, Trans\}$, and when input, these 3DMM coefficients are concatenated as parameters. $ki$ represents latent keypoints in Face-vid2vid. Through MappingNet, we transfer emotional information into facial movement coefficients, thereby generating speaking facial videos with emotional features.

\section{Experiment}

\begin{figure}[!t]\centering
	\includegraphics[width= 0.48\textwidth]{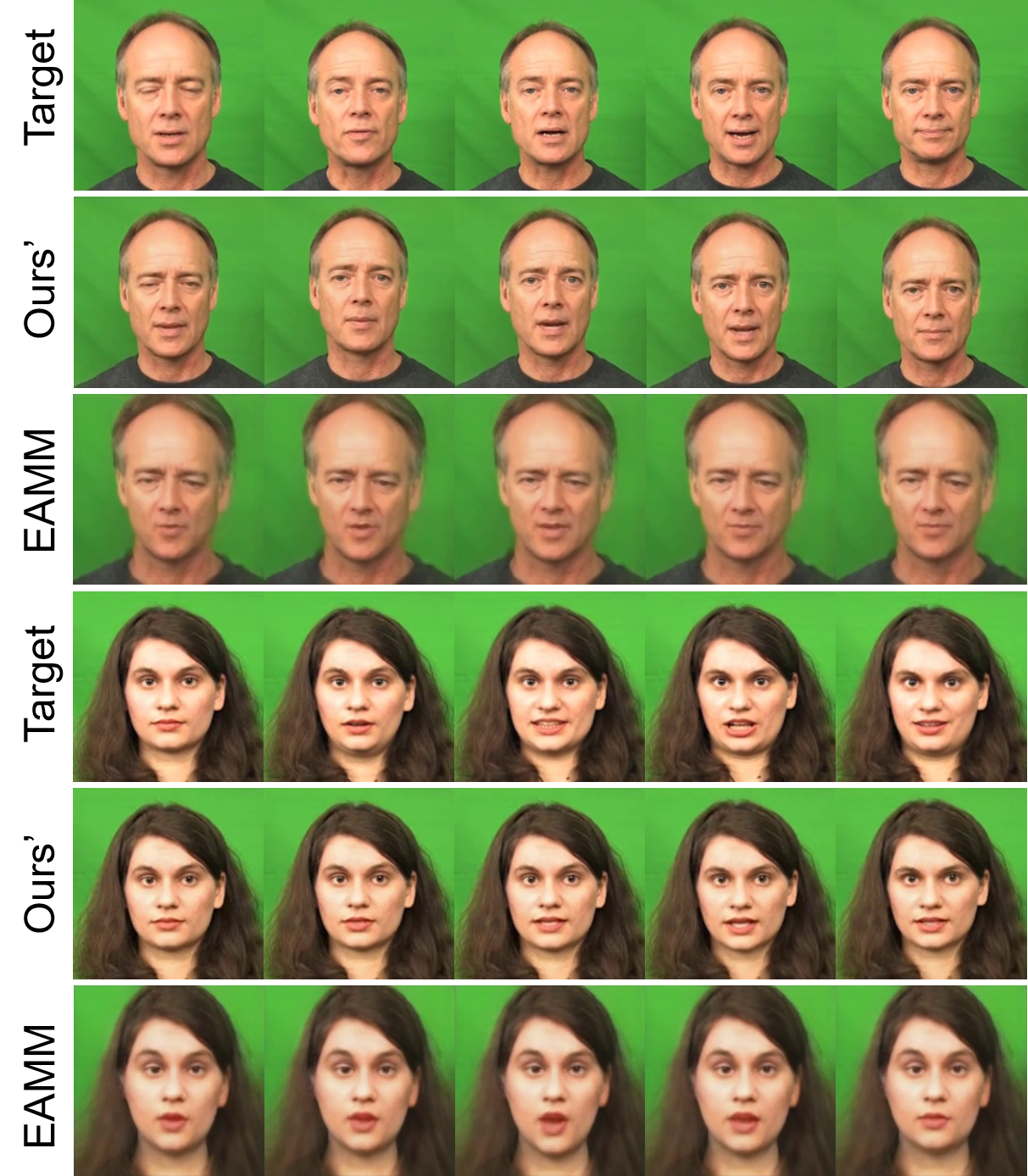}
	\caption{One-shot comparative results on CREAM-D dataset.}
    \label{one-shot}
\end{figure}

\begin{table*}[t]

\caption{Comparison of the results with other SOTA methods. Wav2Lip  generates only the lip region and does not involve the overall generation of the entire face. Therefore, in our comparisons, we do not take into account the results of Wav2Lip. * indicates no comparison. }
\centering
\renewcommand{\arraystretch}{1.2}
% \begin{tabular}{cccccccc}
\begin{tabular}{llllllll}
\hline
\multicolumn{1}{l}{} & \multicolumn{4}{c}{Video Quality} & \multicolumn{3}{c}{Lip Synchronization}           \\ \hline
Method               & FID↓     & SSIM↑ & PSNR↑  & CPBD↑ & Min Dist↓ & AVConf↑ & AV Offset(→0) \\ \hline 

Real Video(MEAD)         & 0.000        & 1.000     & 31.734 & 0.265 & 7.869     & 6.564   & -2.000                          \\ 

% \rowcolor{gray!10}
Wav2lip\cite{prajwal2020lip}              &  16.641$^*$    & 0.932$^*$ & 30.929$^*$  & 0.305$^*$ & \textbf{6.611}     & \textbf{8.119}  & -2.000                          \\ 
% PC-AVS\cite{zhou2021pose}               & 72.598   & 0.305 & 11.387 & 0.088 & 7.751     & 6.876   & -3                          \\ \hline
MEAD\cite{wang2020mead}                 & 146.454  & 0.469 & 14.864 & 0.191 & 11.957    & 2.674   & -2.000                          \\
EVP\cite{ji2021audio}                  & 56.650    & 0.453 & 16.308 & 0.341 & 12.443    & 3.163   & 5.000                           \\
EAMM\cite{ji2022eamm}                 & 204.002  & 0.396 & 12.832 & 0.135 & 10.091    & 3.046   & -4.000                          \\ 

EmoSpeaker(Ours)           & \textbf{25.566} & \textbf{0.728} & \textbf{22.211} & \textbf{0.224} & {8.527} &  {4.354} & \textbf{0.000}                           \\  \hline
Real Video(CREMA-D)      & 0.000        & 1.000     & 26.453 & 0.272 & 7.701     & 6.365   & 0.000                           \\
EAMM\cite{ji2022eamm}        & {113.671} & {0.589} & {13.110} & {0.208} & {10.593} & {3.109} & {2.000}                            \\ 
EmoSpeaker(Ours)            & \textbf{15.418} & \textbf{0.772} & \textbf{23.985} & \textbf{0.289} & \textbf{8.068} & \textbf{4.252} & \textbf{0.000}                            \\ \hline
\end{tabular}
\label{Com_with_SOTA}

\end{table*}

\subsection{Experimental protocols}
% In this section we briefly introduce Dataset, implementation details and evaluation metrics, more details are shown in  supplemental material.
% \subsubsection*{\bf A simple numbered list}
\subsubsection{Dataset}

To achieve our task objectives, we have opted for the MEAD  \cite{wang2020mead} as our training data, aiming to encompass various emotion categories and intensity levels. From this dataset, we have selected speaking videos that cover all emotion categories and intensity levels for 10 males and 10 females, constituting our training set. Additionally, we have chosen 2 males and 2 females as our test set. For One-Shot testing, we have also included the CREMA-D\cite{cao2014crema} dataset and HDTF dataset\cite{zhang2021flow} in our study. In these datasets, we have chosen reference images and driving audio to generate speaking videos with fine-grained emotional expressions.

\subsubsection{Implementation Details}

The experiments are conducted using an NVIDIA RTX 3090Ti GPU on the PyTorch platform and training process is shown in Algorithm \ref{algo}. The input audios are sampled at a rate of 16,000 Hz, and MFCC features are extracted. The input image resolution is set at 512×512 pixels. The input image undergoes processing by OpenFace \cite{baltruvsaitis2016openface} and DeepFace3DReconstruction \cite{deng2019accurate} techniques to obtain AU units and 3DMM coefficients. The Visual Attribute-Guided Audio Decoupler and Fine-grained Emotion Coefficient Prediction Module are trained jointly, whereas the Emotion Face Renderer is trained separately. The total training time required is approximately 30 hours.
The initial learning rate is 1×10-5. We utilize the ADAM optimizer with $\beta_1$ and $\beta_2$ parameters set to 0.9 and 0.999, respectively, and the weight decay is set to 0.001. The training process is halted after 500 epochs. In Visual Attribute-Guided Audio Decoupler, we input 10 frames of images and their corresponding audio simultaneously. However, in  Fine-grained Emotion Coefficient Prediction Module, a sliding window of size 5 is used to train the IdTex coefficient generation, and a sliding window size of 20 is used for the posenet. All other parameters remain the same.
\subsubsection{Evaluation Metrics}

The video quality is evaluated using Frechet Inception Distance (FID) \cite{heusel2017gans}, Structural Similarity (SSIM) \cite{assessment2004error}, Peak Signal to Noise Ratio (PSNR), and Cumulative Probability Blur Detection (CPBD) \cite{narvekar2011no}. Lip synchronization is assessed by using Syncnet \cite{chung2017out} to detect Lip Sync Confidence (AVConf), Lip Offset (AVOff), and Minimum Offset (Min Dist).

\begin{table}[!t]
\caption{User Study results on four differents aspects. The range of scores given by each participant for the video is from 1 to 5. 
 * indicates no comparison.}
\centering
\begin{tabular}{cccccccc}
\cline{1-5}
Method               & \begin{tabular}[c]{@{}c@{}}Lip Syn↑\end{tabular} & \begin{tabular}[c]{@{}c@{}}Emo Acc↑\end{tabular} & \begin{tabular}[c]{@{}c@{}}Video\\ Reality↑\end{tabular} & \begin{tabular}[c]{@{}c@{}}Video\\ Quailty↑\end{tabular} &  &  &  \\ \cline{1-5}
Real Video           & 4.54                                               & 4.74                                                   & 4.88                                                     & 4.89                                                     &  &  &  \\
Wav2Lip              & \textbf{3.96}                                               & 2.51                                                   & *4.40                                                     & *4.65                                                     &  &  &  \\
% PC-AVS               & 3.34                                               & 2.68                                                   & 3.56                                                     & 3.45                                                     &  &  &  \\
MEAD                 & 2.73                                               & 3.42                                                   & 3.87                                                     & 3.72                                                     &  &  &  \\
EVP                  & 2.83                                               & 3.55                                                   & 3.71                                                     & 3.85                                                     &  &  &  \\
EAMM                 & 3.12                                               & 3.72                                                   & 3.65                                                     & 3.56                                                     &  &  &  \\

Ours                 & 3.53                                      & \textbf{3.91}                                          & \textbf{4.21}                                            & \textbf{4.12}                                            &  &  &  \\ \cline{1-5}

\end{tabular}
\label{userstudy}
\end{table}

\begin{table}[!t]
\caption{Lip Synchronization evaluation for Au-based contrastive learning ablation experiments.}

\centering
\begin{tabular}{cccc}
\hline

\multicolumn{1}{l}{} & \multicolumn{3}{c}{Lip Synchronization}           \\ \hline
% \multicolumn{1}{l}{} & \multicolumn{3}{c}{Video Quality}
         & Min Dist↓ & AVConf↑ & AV Offset(→0)\\ \hline

DeepSpeech    &   10.365     &  3.152  &1.000  \\ 
Wav2vec    &   9.991     &  3.374    &2.000  \\ 
without AU &   10.384      &  3.009  &     -1.000    \\
with AU    &   \textbf{8.527}     &  \textbf{4.354}    &\textbf{0.000}  \\ \hline
\end{tabular}
\label{ab_contras}
\end{table}

\begin{figure}[!t]\centering
	\includegraphics[width=7.5cm]{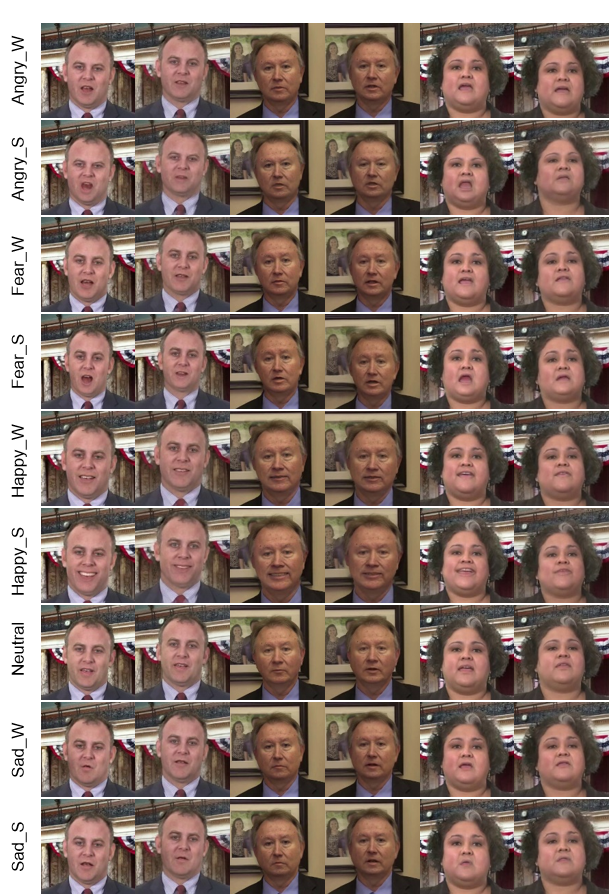}
	\caption{One-shot emotion generation test on HDTF Dataset. 
"W" represents Weak, and "S" represents Strong.}
    \label{one-shot-hdtf}
\end{figure}

\begin{figure}[!t]\centering
	\includegraphics[width= 0.48\textwidth]{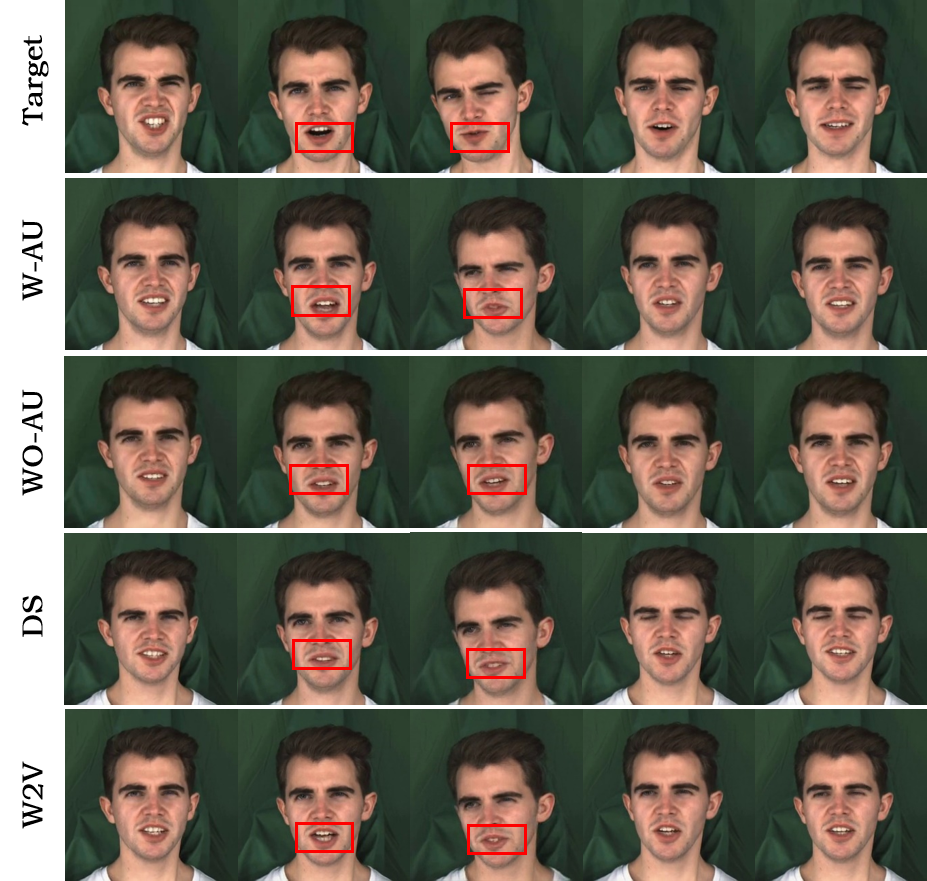}
	\caption{Comparison of video results for AU-based contrastive learning ablation experiment.}
    \label{zui}
\end{figure}

\begin{table*}[!t]
\caption{Quantitative results of mapping net ablation experiments. (T) represent true coefficents and (P) represents coefficient of prediction.}
\centering

\begin{tabular}{cccccccc}
\hline
\multicolumn{1}{l}{} & \multicolumn{4}{c}{Video Quality} & \multicolumn{3}{c}{Lip Synchronization}           \\ \hline
3DMM Combination Mode & FID↓        & SSIM↑      & PSNR↑       & CPBD↑     & Min Dist↓ & AVConf↑ & AV Offset(→0)        \\ \hline
Exp(T) + Id(T)Tex(T)  & 26.313      & 0.717      & \textbf{23.053}      & 0.216     & 9.161              & \textbf{4.582}           & 0.000             \\
Exp(T) + Id(P)Tex(P)  & 27.422      & 0.714      & 22.011      & 0.201     & 9.894              & 4.466           & 0.000             \\
Exp(P) + Id(T)Tex(T)  & 29.149      & 0.725      & 19.809      & 0.215     & 9.372              & 4.487           & -1.000            \\
Exp(P) + Id(P)Tex(P)  & \textbf{25.567}      & \textbf{0.729}      & 22.210      & \textbf{0.224}     & \textbf{8.527}              & 4.354           & \textbf{0.000}             \\ \hline
\end{tabular}
\label{ab_mapping}
\end{table*}

\subsection{Comparison with State-of-the-art Methods}
% \subsubsection{Quantitative evaluation}
% \begin{table*}[t]
% \centering
% %\begin{center}

% 	%\resizebox{.95\columnwidth}{!}{
	
% 	\begin{tabular}{l|cccccc}
% 	   %\hline
% 	   % Evaluation & \multicolumn{6}{c}{\textbf{Sample}}  \\
% 		\hline
% 		Methods & RMSE & MSE & LPIPS & SSIM & FID & PSNR    \\
% 		\hline
% 		MEAD \cite{wang2020mead} & 0.133 & 0.080 & 0.322 & 0.185 & 140.920 & 10.256   \\
% 		Wav2Lip \cite{prajwal2020lip} & 0.112 & 0.045 & 0.162 & 0.296 & 138.925 & 12.182  \\
% 	    PC-AVS \cite{zhou2021pose} & 0.134 & 0.069 &  0.180 & 0.262 & 132.540 & 11.357  \\
	    
% 	    EVP \cite{ji2021audio} & 0.148 & 0.076 & 0.377 & 0.131 & 221.098 & 10.401   \\
% 	      EAMM \cite{ji2022eamm} & 0.148 & 0.076 & 0.377 & 0.131 & 221.098 & 10.401   \\
         
% 	    SadTalker \cite{zhang2023sadtalker} & 0.112 & 0.045 & 0.162 & 0.296 & 138.925 & 12.182  \\
	    
% 	    %Ours(no mask)  & 0.098 & 0.032 & 0.139 & 0.373 & 111.286 & 13.310  \\
	    
% 	    Ours & \textbf{0.096} & \textbf{0.028} & \textbf{0.137} & \textbf{0.378} & \textbf{107.147} & \textbf{14.942}  \\
% 		\hline
% 	\end{tabular}

% %\end{center}
% \caption{Comparison of the results with other SOTA methods.}
% \label{Com_with_SOTA}
% \end{table*}

% \usepackage{multirow}

We compare several state-of-the-art methods for audio-driven emotional facial animation generation (EVP, MEAD, EAMM, etc.) and audio-to-lip generation methods (Wav2Lip) on the MEAD dataset. \textbf{Generating only lip movements is far from sufficient in real-world scenarios. The current approach of the mainstream method, Sadtalker\cite{zhang2023sadtalker}, is not to compare with wav2lip.} Our goal is to accurately control other facial regions while generating lip motion, which can have an impact on lip synchronization. As shown in Table \ref{Com_with_SOTA}, our proposed method exhibits superior performance in overall video quality, and lip synchronization compared to other affective animation generation methods, as quantified by evaluation metrics. We visualize the differences between different methods in the generated results, as shown in the Figure \ref{duibi}. It can be observed that our method has very similar visual quality to the target reference video and can generate corresponding head postures for different affect categories and intensities. In contrast, Ours and Wav2lip can accurately control lips regions. The faces generated by EAMM differ significantly from the reference images, and the facial movements appear unnatural. EVP and MEAD lack accuracy in lip motion and cannot generate fine-grained emotions. In comparison, our proposed method ensures high lip synchronization and richer emotional expression while maintaining video quality.

We conduct one-shot tests on the CREMA-D dataset and compared our results with EAMM. As shown in Figure \ref{one-shot}, we selected multiple reference images and driving audio for video generation. It is evident that our method outperforms EAMM in terms of lip-sync accuracy and exhibits a more natural video quality.

To demonstrate the scalability of our approach, we also conducted one-shot tests on the HDTF dataset. Specifically, the HDTF dataset lacks emotional information, and we select three reference images for emotional video generation. The results in Figure \ref{one-shot-hdtf} indicate that our method successfully controls emotional categories and intensity in the HDTF dataset. It performs well, particularly in generating universally recognized facial expressions such as anger, happiness, and sadness. Additionally, we achieve good results in intensity control.

\subsection{User Study}

To assess the performance of our method compared to current state-of-the-art techniques across various metrics, we create 256 videos with diverse emotions and varying intensities, followed by soliciting ratings from a group of 20 participants. The participants evaluate these videos based on four aspects: Lip Synchronization, Emotion Accuracy, Video Reality, and Video Quality, and the obtained scores are averaged. The scores in Table \ref{userstudy} indicate that Wav2lip excel in lip synchronization, but receive lower scores in Emotional Accuracy and Video Quality as they neglect emotion and employ outdated datasets. Moreover, since Wav2Lip generates only the lip region and does not involve the generation of the entire face, we do not consider the results of Wav2Lip in our comparisons. Conversely, MEAD and EVP outperform in emotional accuracy and video quality as they consider emotion and utilize the newer mead dataset. EAMM yields inferior video quality due to its one-shot generation approach. In contrast, our method attains superior results across all aspects and achieves scores closest to real videos.

\subsection{Ablation Study}
\begin{figure}[!t]\centering
	\includegraphics[width=7.5cm]{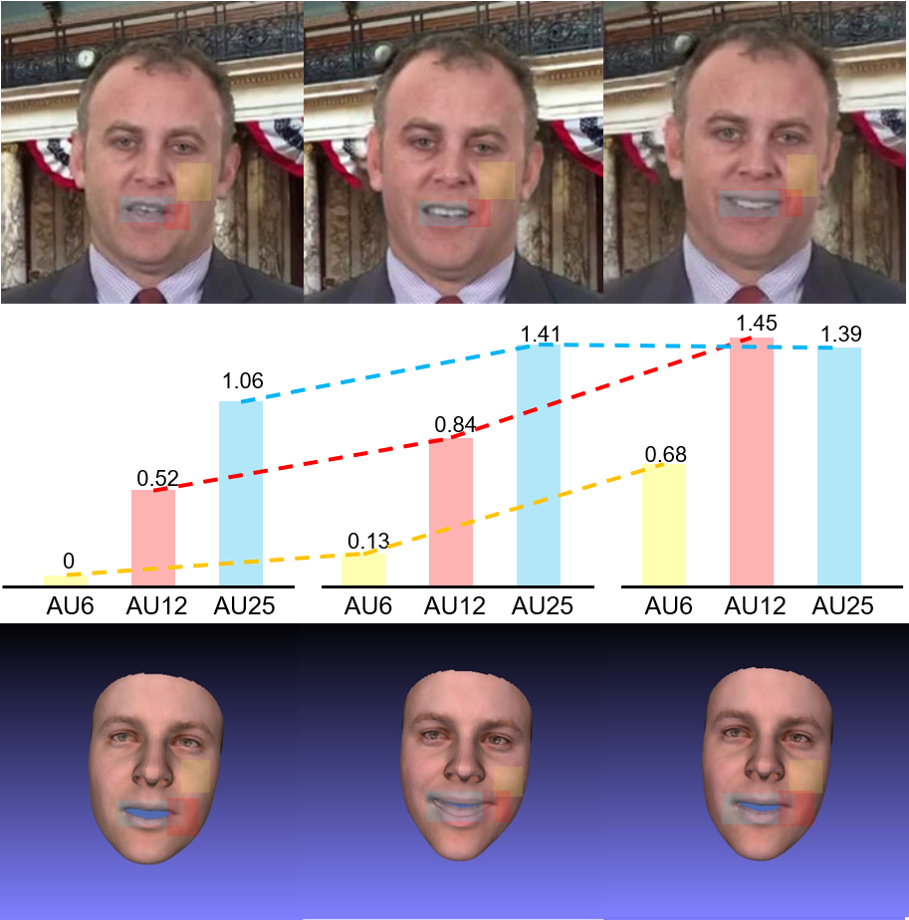}
	\caption{Based on previous research on the correlation between AUs (Action Units) and emotions in Table \ref{tab:au_info} \ref{tab:emotion_au}, it is found that expressing happiness triggers changes in AU6, AU12, and AU25, with higher values indicating greater intensity. We have presented pictures of the same frame from generated videos of different emotional intensities along with their corresponding AU information. The boxed regions indicate the locations of AU movements, and the bar data represent the AU values.}
    \label{auduibi}
\end{figure}

\subsubsection{Ablation of AU-based Contrastive Learning}
In order to investigate the impact of AU-based Contrastive Learning on lip accuracy in Visual Attribute-Guided Audio Decoupler, we compare the results obtained from training the audio encoder using AU-based Contrastive Learning with those without, while keeping other aspects unchanged. The experimental findings in Table \ref{ab_contras} demonstrate that integrating AU-based Contrastive Learning results in a significant improvement of over 30\% in lip movement accuracy for the video speaker. Figure \ref{zui} demonstrates a significant improvement in lip synchronization of the videos generated through AU-based contrastive learning compared to those without. This improvement is primarily evident in the smoothness of lip movements and the accuracy of lip positioning.

We also conduct one-shot emotional generation tests on the HDTF dataset, selecting emotions of varying intensities. For comparison, we utilize AU data and 3D modeling. As shown in  Figure \ref{auduibi}, we choose the emotion of happiness, increasing in intensity from left to right. It can be observed that as the emotional intensity increases, the AUs corresponding to the expressed emotion in the generated videos also increase.
\subsubsection{Ablation of MappingNet}

In Emotion Face Renderer, the final video rendering effect is limited by the pre-training model of Face-vid2vid. In order to more accurately evaluate the effectiveness of our video and exclude the influence of other factors, we conduct the following experiments. The specific method is to divide the lip coefficient exp and the emotion-related coefficients id and tex into two parts, and combine the model's predicted values with the actual extracted values through cross-combination. Finally, the following experimental results in Table \ref{ab_mapping} are obtained. It can be seen that our coefficient prediction model is able to closely approximate the real coefficient values to a large extent, fully demonstrating the effectiveness of our Visual Attribute-Guided Audio Decoupler and Fine-grained Emotion Coefficient Prediction Module.

\subsubsection{Ablation of Fine-grained Emotion Matrix}

\begin{table}[!t]
\caption{Fine-grained emotion intensity quantization matrices.
FL represents inferring sliding window frame length.}
\centering
\begin{tabular}{cccccc}
\hline
Emotion  & FL\_20 & FL\_15 & FL\_10 & FL\_5 & FL\_2 \\ \hline
Happy\_l & 0.749  & 0.747 & 0.804  & 0.884 & 0.979 \\
Happy\_2  &0.864 & 0.903 & 0.932 & 0.937 & 0.977  \\
Happy\_3 & 0.979  & 0.970 & 0.977  & 0.981 & 0.995 \\ \hline
Angry\_l & 0.107  & 0.110 & 0.111  & 0.137 & 0.132 \\
Angry\_2 & 0.144 & 0.225 & 0.241 & 0.241 & 0.281   \\ 
Angry\_3 & 0.133  & 0.278 & 0.289  & 0.299 & 0.320 \\ \hline
Fear\_1  & 0.179  & 0.190 & 0.206  & 0.193 & 0.210 \\
Fear\_2  & 0.178 & 0.179 & 0.203 & 0.218 & 0.304 \\
Fear\_3  & 0.195  & 0.204 & 0.242  & 0.259 & 0.321 \\ \hline
\end{tabular}

\label{ab_fg}
\end{table}

\begin{figure}[!t]\centering
	\includegraphics[width= 0.50\textwidth]{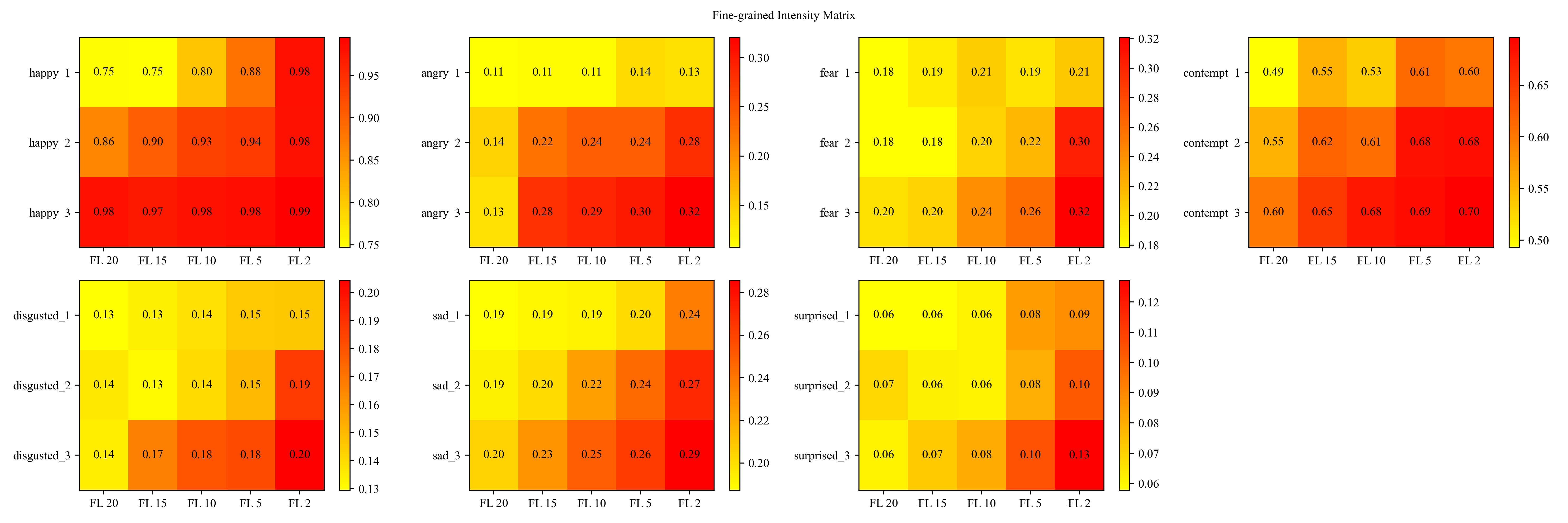}
	\caption{Fine-grained emotion intensity quantization matrice heatmap. FL represents inferring sliding window frame length.
}
    \label{heatmap}
\end{figure}

To examine the impact of fine-grained emotion intensity matrices, we generate videos featuring different emotions by combining varied emotion intensity labels with different audio sliding window sizes. We then classify these generated videos using a pre-trained emotion recognition model \cite{arriaga2020perception}, with the recognition accuracy representing the average probability of correctly classifying all frames within each video. The experiment assumes that a higher recognition probability from the emotion recognition model indicates a stronger intensity of the displayed emotion in the video. The results in Table \ref{ab_fg} demonstrate that altering emotion intensity labels while keeping the inference window size fixed leads to dynamic and substantial changes in emotion intensity. 
Figure \ref{heatmap} demonstrates the effectiveness of our fine-grained emotion control scheme. As the intensity of emotion labels increases and the frame length decreases, the emotional intensity becomes higher. In summary, a fine-grained intensity control matrix, comprising diverse emotion intensity labels and inference window sizes, enables precise control of facial expressions in generated video. \textbf{To more intuitively display the fine-grained effects, we have posted fine-grained videos on the project page instead of presenting them in the form of pictures.}

\section{Conclusion}
This paper presents an algorithm EmoSpeaker for generating emotionally expressive faces with fine-grained intensity, requiring only an audio clip, a portrait, the specified emotions, and intensity granularity. The method extracts content features using a facial emotion decoupling module and incorporates a fine-grained intensity control module to achieve arbitrary emotional intensity.	
This demonstrates promising applications in various fields such as video games, virtual reality, film special effects, and human-computer interface interaction. Subjective and objective evaluations demonstrate the superiority of our method in generating richer facial animations compared to state-of-the-art methods. Future research will focus on conducting in-depth studies in the field of fine-grained intensity control to enhance the generation of more expressive and nuanced facial animations.
\section{Acknowledgement}
The work was jointly supported by the National Key R\&D Program of China under grant No. 2022ZD0117103, the National Natural Science Foundations of China under grant No. 62272364, 62002271, the Teaching Reform Project of Shaanxi Higher Continuing Education under Grant No. 21XJZ004.
\section{Ethical Consideration}
Given that our method enables one-shot facial animation emotion generation, it is easily accessible, thereby raising concerns regarding potential misuse for unlawful activities such as telecommunication fraud. Presently, research efforts are directed towards detecting the authenticity of generated videos. We will openly share the results of our algorithm with the Deepfake \cite{zhao2021multi} community to facilitate the training of detection algorithms and enhance the algorithm's generalization.

\bibliographystyle{ieeetr}
\bibliography{ref}

\begin{thebibliography}{10}

\bibitem{wang2021adversarial}
X.~Wang, W.~Sun, and J.~Jia, ``An adversarial learning framework for high fidelity face completion,'' in {\em Proceedings of the IEEE/CVF Conference on Computer Vision and Pattern Recognition}, pp.~15917--15926, 2021.

\bibitem{yu2020human}
T.~Yu, J.~Wang, and S.-M. Hu, ``Human performance capture using a virtual mirror,'' {\em ACM Transactions on Graphics (TOG)}, vol.~39, no.~4, pp.~1--14, 2020.

\bibitem{zhu2021learning}
L.~Zhu, Y.-K. Tang, and J.~Hays, ``Learning affordance for end-to-end visuomotor robot control in virtual reality,'' in {\em Proceedings of the IEEE/CVF Conference on Computer Vision and Pattern Recognition}, pp.~6995--7004, 2021.

\bibitem{li2020interactive}
B.~Li, J.~Ye, Y.~Wang, and H.~Liu, ``Interactive furniture layout using interior design guidelines,'' {\em ACM Transactions on Graphics (TOG)}, vol.~39, no.~4, pp.~1--14, 2020.

\bibitem{henderson2021digital}
P.~Henderson, Z.~Long, Y.~Zhang, S.~Zhao, and B.~Ghanem, ``Digital actors: A survey of the state of the art,'' {\em ACM Transactions on Graphics (TOG)}, vol.~40, no.~4, pp.~1--21, 2021.

\bibitem{liu2021high}
Y.~Liu, T.~Xiang, X.~Shi, Q.~Dai, and C.~Zhu, ``High-quality facial animation with an adaptive 3d morphable model,'' in {\em IEEE/CVF Conference on Computer Vision and Pattern Recognition (CVPR)}, pp.~7606--7614, IEEE, 2021.

\bibitem{prajwal2020lip}
K.~Prajwal, R.~Mukhopadhyay, V.~P. Namboodiri, and C.~Jawahar, ``A lip sync expert is all you need for speech to lip generation in the wild,'' in {\em Proceedings of the 28th ACM International Conference on Multimedia}, pp.~484--492, 2020.

\bibitem{karras2017audio}
T.~Karras, T.~Aila, S.~Laine, A.~Herva, and J.~Lehtinen, ``Audio-driven facial animation by joint end-to-end learning of pose and emotion,'' {\em ACM Transactions on Graphics (TOG)}, vol.~36, no.~4, pp.~1--12, 2017.

\bibitem{livingstone2018ryerson}
S.~R. Livingstone and F.~A. Russo, ``The ryerson audio-visual database of emotional speech and song (ravdess): A dynamic, multimodal set of facial and vocal expressions in north american english,'' {\em PloS one}, vol.~13, no.~5, p.~e0196391, 2018.

\bibitem{sadoughi2019speech}
N.~Sadoughi and C.~Busso, ``Speech-driven expressive talking lips with conditional sequential generative adversarial networks,'' {\em IEEE Transactions on Affective Computing}, vol.~12, no.~4, pp.~1031--1044, 2019.

\bibitem{wang2020mead}
K.~Wang, Q.~Wu, L.~Song, Z.~Yang, W.~Wu, C.~Qian, R.~He, Y.~Qiao, and C.~C. Loy, ``Mead: A large-scale audio-visual dataset for emotional talking-face generation,'' in {\em Computer Vision--ECCV 2020: 16th European Conference, Glasgow, UK, August 23--28, 2020, Proceedings, Part XXI}, pp.~700--717, Springer, 2020.

\bibitem{ji2021audio}
X.~Ji, H.~Zhou, K.~Wang, W.~Wu, C.~C. Loy, X.~Cao, and F.~Xu, ``Audio-driven emotional video portraits,'' in {\em Proceedings of the IEEE/CVF conference on computer vision and pattern recognition}, pp.~14080--14089, 2021.

\bibitem{ji2022eamm}
X.~Ji, H.~Zhou, K.~Wang, Q.~Wu, W.~Wu, F.~Xu, and X.~Cao, ``Eamm: One-shot emotional talking face via audio-based emotion-aware motion model,'' in {\em ACM SIGGRAPH 2022 Conference Proceedings}, pp.~1--10, 2022.

\bibitem{edwards2016jali}
P.~Edwards, C.~Landreth, E.~Fiume, and K.~Singh, ``Jali: an animator-centric viseme model for expressive lip synchronization,'' {\em ACM Transactions on graphics (TOG)}, vol.~35, no.~4, pp.~1--11, 2016.

\bibitem{li2021write}
L.~Li, S.~Wang, Z.~Zhang, Y.~Ding, Y.~Zheng, X.~Yu, and C.~Fan, ``Write-a-speaker: Text-based emotional and rhythmic talking-head generation,'' in {\em Proceedings of the AAAI Conference on Artificial Intelligence}, vol.~35, pp.~1911--1920, 2021.

\bibitem{wang2021one}
T.-C. Wang, A.~Mallya, and M.-Y. Liu, ``One-shot free-view neural talking-head synthesis for video conferencing,'' in {\em Proceedings of the IEEE/CVF conference on computer vision and pattern recognition}, pp.~10039--10049, 2021.

\bibitem{wang2022one}
S.~Wang, L.~Li, Y.~Ding, and X.~Yu, ``One-shot talking face generation from single-speaker audio-visual correlation learning,'' in {\em Thirty-Sixth {AAAI} Conference on Artificial Intelligence, {AAAI} 2022, Thirty-Fourth Conference on Innovative Applications of Artificial Intelligence, {IAAI} 2022, The Twelveth Symposium on Educational Advances in Artificial Intelligence, {EAAI} 2022 Virtual Event, February 22 - March 1, 2022}, pp.~2531--2539, {AAAI} Press, 2022.

\bibitem{yin2022styleheat}
F.~Yin, Y.~Zhang, X.~Cun, M.~Cao, Y.~Fan, X.~Wang, Q.~Bai, B.~Wu, J.~Wang, and Y.~Yang, ``Styleheat: One-shot high-resolution editable talking face generation via pre-trained stylegan,'' in {\em Computer Vision--ECCV 2022: 17th European Conference, Tel Aviv, Israel, October 23--27, 2022, Proceedings, Part XVII}, pp.~85--101, Springer, 2022.

\bibitem{zhang2023sadtalker}
W.~Zhang, X.~Cun, X.~Wang, Y.~Zhang, X.~Shen, Y.~Guo, Y.~Shan, and F.~Wang, ``Sadtalker: Learning realistic 3d motion coefficients for stylized audio-driven single image talking face animation,'' in {\em Proceedings of the IEEE/CVF Conference on Computer Vision and Pattern Recognition}, pp.~8652--8661, 2023.

\bibitem{ma2023otavatar}
Z.~Ma, X.~Zhu, G.-J. Qi, Z.~Lei, and L.~Zhang, ``Otavatar: One-shot talking face avatar with controllable tri-plane rendering,'' in {\em Proceedings of the IEEE/CVF Conference on Computer Vision and Pattern Recognition}, pp.~16901--16910, 2023.

\bibitem{chen2020simple}
T.~Chen, S.~Kornblith, M.~Norouzi, and G.~Hinton, ``A simple framework for contrastive learning of visual representations,'' in {\em International conference on machine learning}, pp.~1597--1607, PMLR, 2020.

\bibitem{suwajanakorn2017synthesizing}
S.~Suwajanakorn, S.~M. Seitz, and I.~Kemelmacher-Shlizerman, ``Synthesizing obama: learning lip sync from audio,'' {\em ACM Transactions on Graphics (ToG)}, vol.~36, no.~4, pp.~1--13, 2017.

\bibitem{wang2012high}
L.~Wang, W.~Han, and F.~K. Soong, ``High quality lip-sync animation for 3d photo-realistic talking head,'' in {\em 2012 IEEE International Conference on Acoustics, Speech and Signal Processing (ICASSP)}, pp.~4529--4532, IEEE, 2012.

\bibitem{zhang2021facial}
C.~Zhang, Y.~Zhao, Y.~Huang, M.~Zeng, S.~Ni, M.~Budagavi, and X.~Guo, ``Facial: Synthesizing dynamic talking face with implicit attribute learning,'' in {\em Proceedings of the IEEE/CVF international conference on computer vision}, pp.~3867--3876, 2021.

\bibitem{sun2021speech2talking}
Y.~Sun, H.~Zhou, Z.~Liu, and H.~Koike, ``Speech2talking-face: Inferring and driving a face with synchronized audio-visual representation.,'' in {\em IJCAI}, vol.~2, p.~4, 2021.

\bibitem{doukas2021headgan}
M.~C. Doukas, S.~Zafeiriou, and V.~Sharmanska, ``Headgan: One-shot neural head synthesis and editing,'' in {\em Proceedings of the IEEE/CVF International Conference on Computer Vision}, pp.~14398--14407, 2021.

\bibitem{zhou2021pose}
H.~Zhou, Y.~Sun, W.~Wu, C.~C. Loy, X.~Wang, and Z.~Liu, ``Pose-controllable talking face generation by implicitly modularized audio-visual representation,'' in {\em Proceedings of the IEEE/CVF conference on computer vision and pattern recognition}, pp.~4176--4186, 2021.

\bibitem{zhou2020makelttalk}
Y.~Zhou, X.~Han, E.~Shechtman, J.~Echevarria, E.~Kalogerakis, and D.~Li, ``Makelttalk: speaker-aware talking-head animation,'' {\em ACM Transactions On Graphics (TOG)}, vol.~39, no.~6, pp.~1--15, 2020.

\bibitem{DBLP:conf/ijcai/WangLDFY21}
S.~Wang, L.~Li, Y.~Ding, C.~Fan, and X.~Yu, ``Audio2head: Audio-driven one-shot talking-head generation with natural head motion,'' in {\em Proceedings of the Thirtieth International Joint Conference on Artificial Intelligence, {IJCAI} 2021, Virtual Event / Montreal, Canada, 19-27 August 2021} (Z.~Zhou, ed.), pp.~1098--1105, ijcai.org, 2021.

\bibitem{ren2021pirenderer}
Y.~Ren, G.~Li, Y.~Chen, T.~H. Li, and S.~Liu, ``Pirenderer: Controllable portrait image generation via semantic neural rendering,'' in {\em Proceedings of the IEEE/CVF International Conference on Computer Vision}, pp.~13759--13768, 2021.

\bibitem{zhang2021flow}
Z.~Zhang, L.~Li, Y.~Ding, and C.~Fan, ``Flow-guided one-shot talking face generation with a high-resolution audio-visual dataset,'' in {\em Proceedings of the IEEE/CVF Conference on Computer Vision and Pattern Recognition}, pp.~3661--3670, 2021.

\bibitem{sinhaemotion2022ijcai}
S.~Sinha, S.~Biswas, R.~Yadav, and B.~Bhowmick, ``Emotion-controllable generalized talking face generation,'' 2022.

\bibitem{DBLP:conf/nips/Wang0TLCK19}
T.~Wang, M.~Liu, A.~Tao, G.~Liu, B.~Catanzaro, and J.~Kautz, ``Few-shot video-to-video synthesis,'' in {\em Advances in Neural Information Processing Systems 32: Annual Conference on Neural Information Processing Systems 2019, NeurIPS 2019, December 8-14, 2019, Vancouver, BC, Canada} (H.~M. Wallach, H.~Larochelle, A.~Beygelzimer, F.~d'Alch{\'{e}}{-}Buc, E.~B. Fox, and R.~Garnett, eds.), pp.~5014--5025, 2019.

\bibitem{zakharov2020fast}
E.~Zakharov, A.~Ivakhnenko, A.~Shysheya, and V.~Lempitsky, ``Fast bi-layer neural synthesis of one-shot realistic head avatars,'' in {\em Computer Vision--ECCV 2020: 16th European Conference, Glasgow, UK, August 23--28, 2020, Proceedings, Part XII 16}, pp.~524--540, Springer, 2020.

\bibitem{siarohin2019first}
A.~Siarohin, S.~Lathuili{\`e}re, S.~Tulyakov, E.~Ricci, and N.~Sebe, ``First order motion model for image animation,'' {\em Advances in Neural Information Processing Systems}, vol.~32, 2019.

\bibitem{siarohin2019animating}
A.~Siarohin, S.~Lathuili{\`e}re, S.~Tulyakov, E.~Ricci, and N.~Sebe, ``Animating arbitrary objects via deep motion transfer,'' in {\em Proceedings of the IEEE/CVF Conference on Computer Vision and Pattern Recognition}, pp.~2377--2386, 2019.

\bibitem{burkov2020neural}
E.~Burkov, I.~Pasechnik, A.~Grigorev, and V.~Lempitsky, ``Neural head reenactment with latent pose descriptors,'' in {\em Proceedings of the IEEE/CVF conference on computer vision and pattern recognition}, pp.~13786--13795, 2020.

\bibitem{blanz1999morphable}
V.~Blanz and T.~Vetter, ``A morphable model for the synthesis of 3d faces,'' in {\em Proceedings of the 26th annual conference on Computer graphics and interactive techniques}, pp.~187--194, 1999.

\bibitem{ekman1978facial}
P.~Ekman and W.~V. Friesen, ``Facial action coding system,'' {\em Environmental Psychology \& Nonverbal Behavior}, 1978.

\bibitem{amodei2016deep}
D.~Amodei, S.~Ananthanarayanan, R.~Anubhai, J.~Bai, E.~Battenberg, C.~Case, J.~Casper, B.~Catanzaro, Q.~Cheng, G.~Chen, {\em et~al.}, ``Deep speech 2: End-to-end speech recognition in english and mandarin,'' in {\em International conference on machine learning}, pp.~173--182, PMLR, 2016.

\bibitem{baevski2020wav2vec}
A.~Baevski, Y.~Zhou, A.~Mohamed, and M.~Auli, ``wav2vec 2.0: A framework for self-supervised learning of speech representations,'' {\em Advances in neural information processing systems}, vol.~33, pp.~12449--12460, 2020.

\bibitem{cao2014crema}
H.~Cao, D.~G. Cooper, M.~K. Keutmann, R.~C. Gur, A.~Nenkova, and R.~Verma, ``Crema-d: Crowd-sourced emotional multimodal actors dataset,'' {\em IEEE transactions on affective computing}, vol.~5, no.~4, pp.~377--390, 2014.

\bibitem{baltruvsaitis2016openface}
T.~Baltru{\v{s}}aitis, P.~Robinson, and L.-P. Morency, ``Openface: an open source facial behavior analysis toolkit,'' in {\em 2016 IEEE winter conference on applications of computer vision (WACV)}, pp.~1--10, IEEE, 2016.

\bibitem{deng2019accurate}
Y.~Deng, J.~Yang, S.~Xu, D.~Chen, Y.~Jia, and X.~Tong, ``Accurate 3d face reconstruction with weakly-supervised learning: From single image to image set,'' in {\em IEEE Computer Vision and Pattern Recognition Workshops}, 2019.

\bibitem{heusel2017gans}
M.~Heusel, H.~Ramsauer, T.~Unterthiner, B.~Nessler, and S.~Hochreiter, ``Gans trained by a two time-scale update rule converge to a local nash equilibrium,'' {\em Advances in neural information processing systems}, vol.~30, 2017.

\bibitem{assessment2004error}
I.~Q. Assessment, ``From error visibility to structural similarity,'' {\em IEEE transactions on image processing}, vol.~13, no.~4, p.~93, 2004.

\bibitem{narvekar2011no}
N.~D. Narvekar and L.~J. Karam, ``A no-reference image blur metric based on the cumulative probability of blur detection (cpbd),'' {\em IEEE Transactions on Image Processing}, vol.~20, no.~9, pp.~2678--2683, 2011.

\bibitem{chung2017out}
J.~S. Chung and A.~Zisserman, ``Out of time: automated lip sync in the wild,'' in {\em Computer Vision--ACCV 2016 Workshops: ACCV 2016 International Workshops, Taipei, Taiwan, November 20-24, 2016, Revised Selected Papers, Part II 13}, pp.~251--263, Springer, 2017.

\bibitem{arriaga2020perception}
O.~Arriaga, M.~Valdenegro-Toro, M.~Muthuraja, S.~Devaramani, and F.~Kirchner, ``Perception for autonomous systems (paz),'' 2020.

\bibitem{zhao2021multi}
H.~Zhao, W.~Zhou, D.~Chen, T.~Wei, W.~Zhang, and N.~Yu, ``Multi-attentional deepfake detection,'' in {\em Proceedings of the IEEE/CVF conference on computer vision and pattern recognition}, pp.~2185--2194, 2021.

\end{thebibliography}

\begin{IEEEbiography}[{\includegraphics[width=1in,height=1.25in,clip,keepaspectratio]{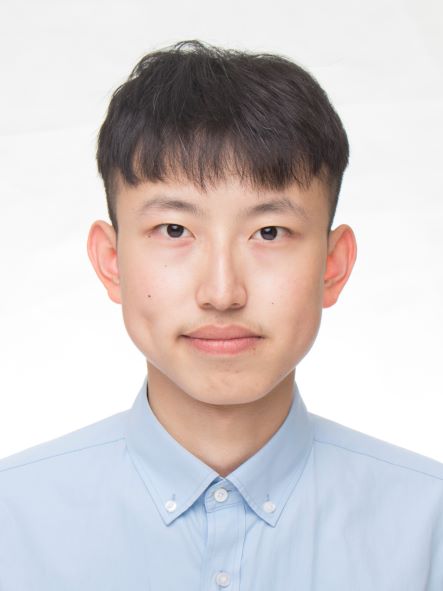}}]{Guanwen Feng}
received the B.S. degree in software
engineering from Hangzhou Dianzi University in
2020. He is current a Ph.D. candidate with School
of Computer Science and Technology, Xidian University.
His rescarch interests include  talking face animation, sign language generation, traffic prediction.
\end{IEEEbiography}

\begin{IEEEbiography}[{\includegraphics[width=1in,height=1.25in,clip,keepaspectratio]{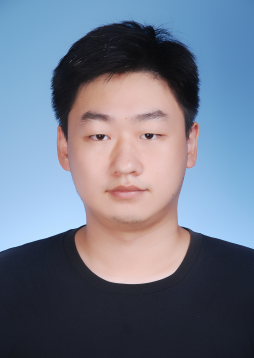}}]{Haoran Cheng}
received the B.E. degree in computer science and technology from Hainan University in 2021. He is current a master degree candidate with School of Computer Science and Technology, Xidian University. His research interests include talking face generation and computer vision.
\end{IEEEbiography}

\begin{IEEEbiography}[{\includegraphics[width=1in,height=1.25in,clip,keepaspectratio]{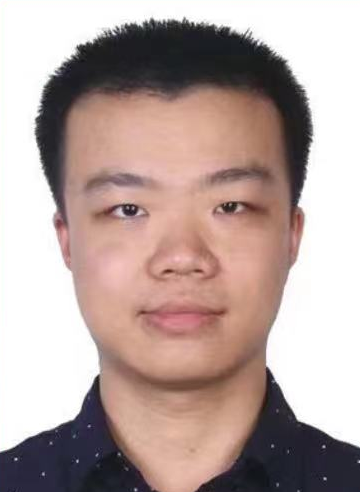}}]{Yunan Li}
received his B.S and Ph.D. degree from the School of Computer Science and Technology, Xidian University, Xi’an, China in 2014 and 2019, respectively. He is current a Huashan Elite Associate Professor in Xidian University. His research interests include computer vision and pattern recognition, especially their applications in image enhancement and action/gesture recognition.  
\end{IEEEbiography}

\begin{IEEEbiography}[{\includegraphics[width=1in,height=1.25in,clip,keepaspectratio]{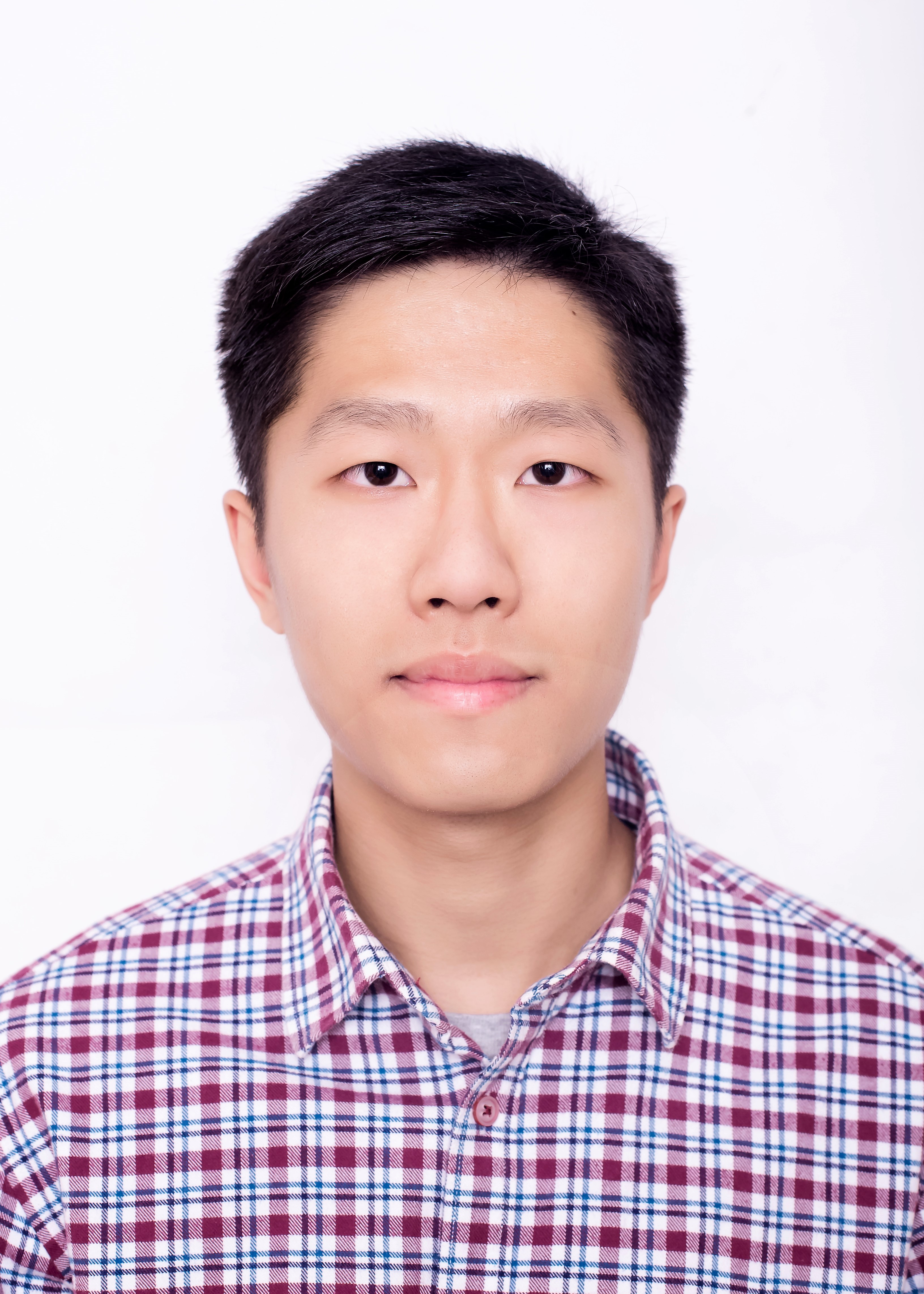}}]{Zhiyuan Ma}
received his B.S. degree in computer science and technology from Zhejiang Gongshang University in 2021. He is currently a M.S. candidate at the School of Computer Science and Technology, Xidian University. His research interests include talking face generation and object detection.
\end{IEEEbiography}

\begin{IEEEbiography}[{\includegraphics[width=1in,height=1.25in,clip,keepaspectratio]{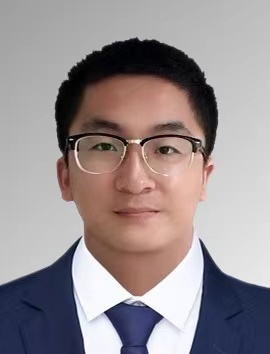}}]{Chaoneng Li}
is a Ph.D. candidate with School of Computer Science and Technology, Xidian University. Prior to that, he received his B.S. degree in the Internet of Things in 2017 from Northwest Normal University. Recently, his main research focuses on computer vision, anomaly detection, trajectory mining and traffic prediction.
\end{IEEEbiography}

\begin{IEEEbiography}[{\includegraphics[width=1in,height=1.25in,clip,keepaspectratio]{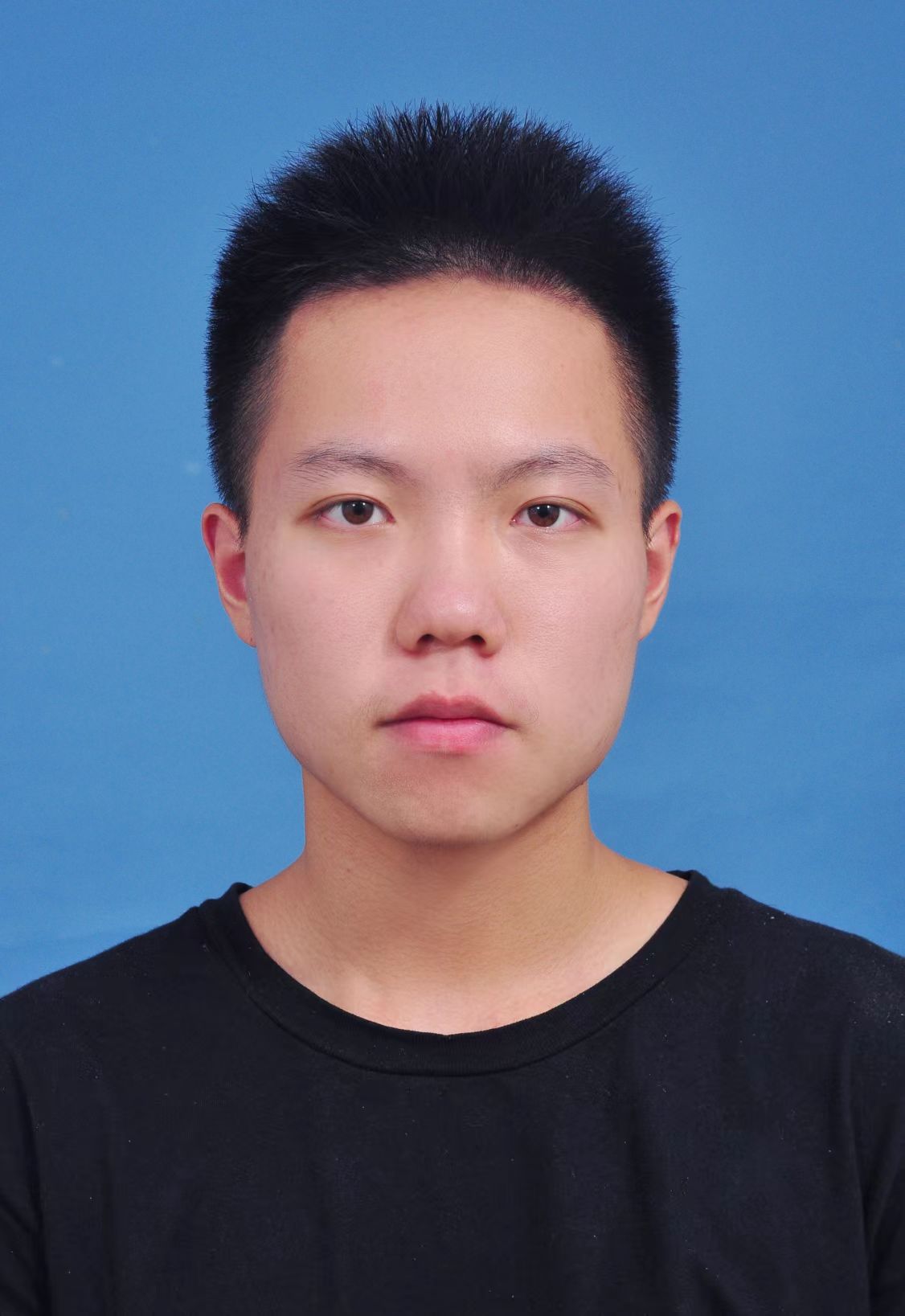}}]{Zhihao Qian}
is an undergraduate at the School of Computer Science and Technology, Xidian University, majoring in Computer Science and Technology. Recently, his main research focus has been on computer vision, notably facial reenactment.
\end{IEEEbiography}

\begin{IEEEbiography}[{\includegraphics[width=1in,height=1.25in,clip,keepaspectratio]{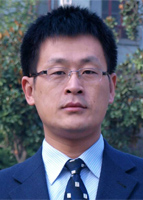}}]{Qiguang Miao}
 is the professor and Ph.D. student supervisor of School of Computer Science and Technology in Xidian University. He received his Ph.D. degree from Xidian University in 2005. His research interests include intelligent image/video understanding and big data. In recent years, He has published over 100 papers in the significant international journals or conferences.
\end{IEEEbiography}

\begin{IEEEbiography}[{\includegraphics[width=1in,height=1.25in,clip,keepaspectratio]{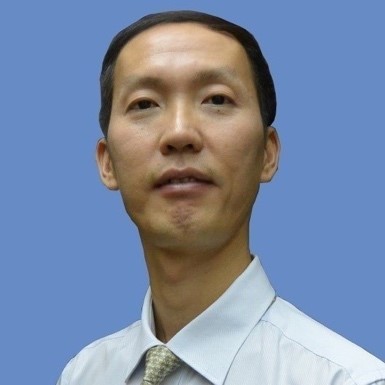}}]{Chi-Man Pun}
received his Ph.D. degree in Computer Science and Engineering from the Chinese University of Hong Kong in 2002, and his M.Sc. and B.Sc. degrees from the University of Macau. He had served as the Head of the Department of Computer and Information Science, University of Macau from 2014 to 2019, where he is currently a Professor and in charge of the Image Processing and Pattern Recognition Laboratory. He has investigated many externally funded research projects as PI, and has authored/co-authored more than 200 refereed papers in many top-tier journals and conferences. He has also co-invented several China/US Patents, and is the recipient of the Macao Science and Technology Award 2014. Dr. Pun has served as the General Chair/Co-chair and the Program/Local Chair for many international conferences. He has also served as the SPC/PC member for many top CS conferences such as AAAI, CVPR, ICCV, ECCV, MM, etc. He has been listed in the World's Top 2\% Scientists by Stanford University since 2020. His research interests include Image Processing and Pattern Recognition; Multimedia Information Security, Forensic and Privacy; Adversarial Machine Learning and AI Security, etc.
\end{IEEEbiography}
\vfill

\end{document}